# Hard-Stop Synthesis for Multi-DOF Compliant Mechanisms

Dean Chen[a], Armin Pomeroy[a], Brandon T. Peterson[a], Will Flanagan[a], He Kai Lim[a], Alexandra Stavrakis[b], Nelson F. SooHoo[b], Jonathan B. Hopkins[a] and Tyler R. Clites[a],[*]

[a] *University of California, Los Angeles, Department of Mechanical and Aerospace Engineering, 420 Westwood Plaza, Los Angeles, CA 90095, USA*
[b] *University of California, Los Angeles, David Geffen School of Medicine., 10833 Le Conte Ave, Los Angeles, CA 90095, USA*
[*] *Corresponding author. E-mail address: clites@ucla.edu (T. R. CLITES).*

**Abstract**

Compliant mechanisms have significant potential in precision applications due to their ability to guide motion without contact. However, an inherent vulnerability to fatigue and mechanical failure has hindered the translation of compliant mechanisms to real-world applications. This is particularly challenging in service environments where loading is complex and uncertain, and the cost of failure is high. In such cases, mechanical hard stops are critical to prevent yielding and buckling. Conventional hard-stop designs, which rely on stacking single-DOF limits, must be overly restrictive in multi-DOF space to guarantee safety in the presence of unknown loads. In this study, we present a systematic design synthesis method to guarantee overload protection in compliant mechanisms by integrating coupled multi-DOF motion limits within a single pair of compact hard-stop surfaces. Specifically, we introduce a theoretical and practical framework for optimizing the contact surface geometry to maximize the mechanism's multi-DOF working space while still ensuring that the mechanism remains within its elastic regime. We apply this synthesis method to a case study of a caged-hinge mechanism for orthopaedic implants, and provide numerical and experimental validation that the derived design offers reliable protection against fatigue, yielding, and buckling. This work establishes a foundation for precision hard-stop design in compliant systems operating under uncertain loads, which is a crucial step toward enabling the application of compliant mechanisms in real-world systems.

1. Introduction

Compliant mechanisms have transformative potential as bearing elements due to their ability to guide motion through elastic deformation rather than relying on traditional sliding or rolling joints [1–14]. When designing compliant mechanisms for specific machine applications, the bulk of design efforts typically focus on flexure arrangement and topology [8,9,15], deformation sensing [16,17], and actuation [18,19]. In practice, however, performance of compliant systems—particularly in terms of range of motion, structural strength, and stability—is fundamentally limited by the risk of mechanical failure (yielding and buckling). This is of particular concern in applications where the compliant mechanism is exposed to unpredictable complex three-dimensional (3D) loads; in such settings, it is critical to provide protection from overloading that could otherwise induce plastic deformation in the mechanism. Without such protections, the compliant bearing could fail catastrophically if exposed to unexpected loads.

One means of protection for compliant mechanisms is to introduce contact-based structures [20–22] that help limit motion of the stage relative to the ground. The two most common approaches to this are contact-aided flexures (CAF) [23–26] and mechanical hard stops [27,28]. CAFs are designed to guide mechanism deformation under known, cyclic loads rather than unpredictable, multi-DOF overloads. In this way, CAFs typically make contact each time the mechanism is loaded; this repeated contact accelerates wear, which is particularly problematic in orthopaedic applications. In contrast, hard stops serve as a motion-limiting safeguard, designed exclusively for highly uncertain, multi-DOF overloads: the hard stop remains untouched during normal function, and engages instantaneously in the presence of supra-threshold loads according to precisely-designed kinematic boundaries. More importantly, CAFs themselves are inherently flexible, and would therefore still rely on hard stops for overload protection [29]. While flexible stops [30–33] have proven effective for managing predictable internal loads by absorbing energy [34,35] rather than enforcing rigid constraints, they fall short under heavy external uncertainties. Fundamentally, flexible stops are designed for "energy-determinant" loads whereas the hard stops are constructed for "time-variant" loads – which are not consumable and may lead to unlimited energy input if given no rigid constraint.

In contrast to these other approaches, rigid hard stops provide reliable, non-energy-limited protection for mechanisms subjected to unpredictable external loads. Although implementation of single-DOF hard stops is straightforward [28,36], the design complexity increases dramatically for loads that span multiple of the six DOFs in the Cartesian force and moment space. In fact, a true safety guarantee requires hard limits in every possible load direction—not just within the mechanism's intended degrees of freedom—since compliant mechanisms can fail under overload or instability in any direction. In fact, many mechanisms are most susceptible to failure in loading directions other than their primary motion DOF (e.g. a cross-axis flexural pivot in compression [37,38]). In a conventional protection scheme, single-DOF hard stops are designed independently for each motion axis; however, we will show that this intersection of single-DOF motion spaces inevitably results in overprotection and unnecessary reduction of the mechanism's working space.

As an alternative to stacked single-DOF hard stops, we introduce multi-DOF hard stops that couple motion limits simultaneously across different load and movement DOFs. We accomplish this through carefully engineered contact *surfaces* that engage under overload conditions to guide motion of the mechanism in 3D space, only allowing motion in directions that maintain stress below a predetermined threshold (e.g. fatigue limit, yielding, and buckling). Designing such systems is considerably more challenging than implementing single-DOF hard stops, because it requires 3D optimization of the contact surface, often within a compact geometric envelope. In this manuscript, we present a general framework for the design of multi-DOF hard stops, involving (1) analysis of the mechanism's stress response, (2) selection of an appropriate hard-stop configuration, and then (3) optimization of the parameters that govern the hard stop's performance. We then apply this general framework to optimization of hard stops for a representative case study in the field of orthopaedic implants [37,39–43]. This is an ideal initial case study because there are high risks of overload due to the unpredictable nature of physiological loading during complex activities or accidental impacts [44–47], and the cost of failure is large enough to require a robust guarantee of protection from overloading. In the context of a compliant stem for total knee arthroplasty (TKA), we demonstrate in both simulation and on the benchtop our framework's ability to produce a hard stop that prioritizes (1) achieving a 100-year fatigue life under physiological loading [40,48–50] and (2) minimizing hard-stop contact frequency to reduce wear and potential patient discomfort.

## 2. Hard-Stop Design Synthesis
### 2.1 Design objectives for hard stops

We begin with a general framework for synthesizing hard-stop designs to protect multiple degree-of-freedom (DOF) compliant mechanisms. First, we consider a general compliant mechanism in which one end is fixed (ground end $G$), while the other end is free to move (namely, free-stage end $A$). The free end is subjected to an $S$-DOF load, expressed as

$$W = \sum_{i=1}^{S} w_i d_i, \tag{1}$$

where $d_i$ denotes the unit vector of each degree of freedom and $w_i$ is the corresponding load magnitude (for example, force or moment). $S$ denotes the total number of DOFs available in the global space; for instance, in a typical three-dimensional Cartesian space, $S = 6$ (three force and three moment components). We assume that the applied load induces a rigid-body motion of the free end $A$ as a result of the mechanism's elastic deformation, and represent this motion by

$$U_S = \sum_{i=1}^{S} u_i f_i. \tag{2}$$

Here, the motion vector $U_S$ is defined in the coordinate system $\{f_i\}$, which may either be the same as the load's coordinate system $\{d_i\}$ or a transform of it—either through a linear transformation or a nonlinear transformation between coordinate systems, such as cylindrical, Cartesian, or spherical. The mapping relation $U_S = k(W)$ is referred to as the "compliance function" of the compliant system. A key characteristic of compliant systems is their direction-dependent stiffness. Conventionally, a binary classification is considered: number $K$ of compliant DOFs and $(S - K)$ rigid DOFs. However, under highly uncertain loads, no specific DOF can be rigorously treated as rigid; in these cases, the typical binary classification becomes inadequate. As such, in the context of overload protection, we select DOFs for hard-stop design based not on the *compliance* in each DOF, but on whether that DOF needs to be protected. In other words, a DOF should be included in the hard-stop design if deformation in that DOF could conceivably contribute to system failure—such as yielding, fatigue, or instability—under loads that are *possible*. In this study, we define $K$ as the total number of DOFs that require protection, while the remaining number $S - K$ are those that do not necessitate protection. The motion vector corresponding to the protected DOFs is then defined as

$$\boldsymbol{U} = \sum_{i=1}^{K} u_i \boldsymbol{e}_i, \tag{3}$$

where the unit vector set $\{\boldsymbol{e}_i\} \subseteq \{\boldsymbol{f}_i\}$ is referred to as the "*workspace*" of the hard stop. Note that this workspace is typically selected based on the convenience of physical construction or shape optimization of the hard stops, which may not align with the coordinate system $\{\boldsymbol{d}_i\}$ of external loads.

In engineering applications, various criteria are used to determine whether the mechanism operates within a safe range, such as maximum stress, strain limits, or system stability. Here, we adopt the most commonly used criterion: the maximum stress limit $\sigma^{cr}$, which may correspond to yielding, fatigue endurance, or similar stress thresholds. Next, we consider that the elastic deformation of the internal mechanism induces a specific internal stress distribution, where the maximum stress in the field is defined as $\sigma^{max}$. For a non-path-dependent, purely elastic system, we assume a bijective relationship between the motion of the free stage and the maximum internal stress, expressed as

$$\sigma^{max} = \mathbb{R}(\boldsymbol{U}), \tag{4}$$

and referred to as the "stress response function".

Therefore, the macroscopic behavior of the compliant mechanism under input loads can be described as the relation sequence: $\boldsymbol{W} \to \boldsymbol{U} \to \sigma^{max}$. We consider that the safe operation of the compliant mechanism requires $\sigma^{max} < \sigma^{cr}$. Then, the set of threshold motion vectors that lead to critical internal stresses can be defined as:

$$\{\boldsymbol{U}\}^{cr} = \{\boldsymbol{U} : \mathbb{R}(\boldsymbol{U}) = \sigma^{cr}\}. \tag{5}$$

Consider a finite, closed boundary surface $\boldsymbol{\Gamma}^\sigma$, defined by the set of critical deformation vectors $\{\boldsymbol{U}\}^{cr}$ in the $K$-dimensional space, namely,

$$\boldsymbol{\Gamma}^\sigma = \{\boldsymbol{U}\}^{cr}. \tag{6}$$

The finite volume enclosed by the closed surface $\boldsymbol{\Gamma}^\sigma$ is defined as $\boldsymbol{\mathcal{R}}_\sigma$, which we will refer to as the "safe stress space". By definition, any motion vector $\boldsymbol{U} \in \boldsymbol{\mathcal{R}}_\sigma$ will not lead to the internal stress of the mechanism exceeding the stress limit $\sigma^{cr}$.

Note that we assume a conservative stress response, namely, $\boldsymbol{\mathcal{R}}_\sigma$ is a radially convex (monotonic) space. Specifically, for any motion vector $\boldsymbol{U}$, the condition $\mathbb{R}(k\boldsymbol{U}) < \mathbb{R}(\boldsymbol{U})$ holds for any $0 < k < 1$. This means that reducing the magnitude of $\boldsymbol{U}$ along the same direction will *monotonically* (but not necessarily proportionally) reduce the internal stress within the mechanism. The safe stress space $\boldsymbol{\mathcal{R}}_\sigma$ can be derived once $\boldsymbol{\Gamma}^\sigma$ is determined,

$$\boldsymbol{\mathcal{R}}_\sigma = \{\boldsymbol{U} : \boldsymbol{U} = k\boldsymbol{U}^{cr}, \boldsymbol{U}^{cr} \in \boldsymbol{\Gamma}^\sigma, 0 < k < 1\}. \tag{7}$$

Next, we define another $K$-dimensional volume, $\mathcal{R}_{hs}$, as the "hard-stop-free workspace". Specifically, for any $U \in \mathcal{R}_{hs}$, the motion of the free-stage end $A$ will not result in any contact engagement. Therefore, a strict formulation of the primary objective in hard-stop design is to ensure that the hard-stop-free workspace is fully contained within the safe stress space, namely,

$$\mathcal{R}_{hs} \subseteq \mathcal{R}_\sigma. \tag{8}$$

When Eq. (8) is not strictly satisfied, there will exist an "unprotected motion space", $\mathcal{R}_{hs\setminus\sigma}$, which is defined as the portion of the hard-stop-free workspace that lies outside the safe stress space:

$$\mathcal{R}_{hs\setminus\sigma} = \{U: U \in \mathcal{R}_{hs}, U \notin \mathcal{R}_\sigma\}, \text{if } \mathcal{R}_{hs\setminus\sigma} \neq \emptyset. \tag{9}$$

Any motion within $\mathcal{R}_{hs\setminus\sigma}$ results in internal stress exceeding the allowable limit, without triggering hard-stop engagement, which exposes the mechanism to potential failure.

When Eq. (8) *is* strictly satisfied, we introduce a secondary design criterion: the volume fraction of the hard-stop-free workspace within the safe stress space, defined as

$$\phi_{hs} = \frac{Vol(\mathcal{R}_{hs})}{Vol(\mathcal{R}_\sigma)}. \tag{10}$$

If $\phi_{hs} < 1$, there exists an "overprotected motion space", $\mathcal{R}_{\sigma\setminus hs}$, defined as the portion of the safe stress space unnecessarily limited by hard stops, namely,

$$\mathcal{R}_{\sigma\setminus hs} = \{U: U \notin \mathcal{R}_{hs}, U \in \mathcal{R}_\sigma\}, \text{if } \mathcal{R}_{\sigma\setminus hs} \neq \emptyset. \tag{11}$$

Any motion $U \in \mathcal{R}_{\sigma\setminus hs}$ does *not* cause internal stress to exceed the limit, and yet the hard stop would already engage and prevent this motion. This leads to unnecessary hard-stop wear, impact, and sacrifice in working space of the compliant mechanism. Therefore, the secondary objective in hard-stop design is to maximize $\phi_{hs}$, thereby minimizing the frequency of unnecessary contact and wear during lower-extremity tasks. This optimization goal can be expressed as

$$\max \phi_{hs}, \tag{12}$$

subject to Eq. (8).

2.2 Stress response function

As discussed above, the effective design of hard stops requires a precisely optimized balance between ensuring "sufficient protection" (Eq. (8)) and avoiding "overprotection" (Eq. (12)). The difficulty of achieving an optimal balance of these priorities largely depends on the shape of the stress response space $\mathcal{R}_\sigma$, while $\mathcal{R}_{hs}$ is typically optimized as a maximum inscribed volume within $\mathcal{R}_\sigma$. The maximum achievable volume fraction of this inscribed region depends on the geometry of $\mathcal{R}_\sigma$ and $\mathcal{R}_{hs}$; for example, the maximum volume fraction of an inscribed rectangle differs in triangles, rectangles, and circles.

We begin by examining a key property of the stress response space: is $\mathcal{R}_\sigma$ orthogonal or skewed in the $\{e_i\}$ coordinate space? If $\mathcal{R}_\sigma$ is an orthotope, on the critical boundary $\Gamma^\sigma$, the stress response in a specific degree of freedom is influenced solely by motion in that direction. To explore the question of skew, we examine several examples of simple compliant systems. First, we consider a specific compliant mechanism in which the stress response is governed by a simple linear superposition across all degrees of freedom, namely:

$$\sigma^{max} = \mathbb{R}(\boldsymbol{U}) = \sum_{i=1}^{K} R_i |u_i|, \tag{13}$$

where $\{R_i\}$ are a set of constants. It can be derived that, even for this linear system, the shape of $\mathcal{R}_\sigma$ is not an orthotope. For example, in a two-DOF case of Eq. (13) (Fig. 1a), the deflection ($u_1$) and axial displacement ($u_2$) of a cantilever cylinder beam results in a superposition of the system's maximum principal stress. Here, $\boldsymbol{U} = u_1 \boldsymbol{e}_1 + u_2 \boldsymbol{e}_2$, and $\sigma^{max} = R_1|u_1| + R_2|u_2|$ (under small-deformation assumptions). The critical boundary $\boldsymbol{\Gamma}^\sigma$, obtained via Eqs. (6) and (7) forms a diamond-shaped region connecting the points $(\sigma^{cr}/R_1, 0)$, $(0, \sigma^{cr}/R_2)$, $(-\sigma^{cr}/R_1, 0)$, and $(0, -\sigma^{cr}/R_2)$. It is important to note that a linear transformation of the coordinate system still results in a diamond shape $\mathcal{R}_\sigma$.

Second, we consider a different stress response determined by the magnitude (radius) of the motion vector $\boldsymbol{U}$, namely,

$$\sigma^{max} = \mathbb{R}(\boldsymbol{U}) = R' \left[ \sum_{i=1}^{K} u_i^2 \right]^{1/2}. \tag{14}$$

A representative two-DOF example is the lateral translation of a cantilever cylindrical beam in two in-plane directions, where $\sigma^{max} = R'(u_1^2 + u_2^2)^{1/2}$. In this case, $\mathcal{R}_\sigma$ takes the form of a circular region, as illustrated in Fig. 1b.

Stress response may also exhibit synergistic or antagonistic effects between motions in different directions. As an example, we consider the coupling between angular and translational deflection in a cylindrical beam (Fig. 1c). The resulting $\mathcal{R}_\sigma$ for such a system resembles an inclined "rugby ball" shape. These special cases are further examined through numerical simulations in Appendix. A and Sec. 3.3. In summary, we observe that $\mathcal{R}_\sigma$ is not an orthotope in most compliant systems. Therefore, finding the design of hard stops that can well maximum $\phi_{hs}$ in skewed stress response space is critical for the protection of compliant mechanism.

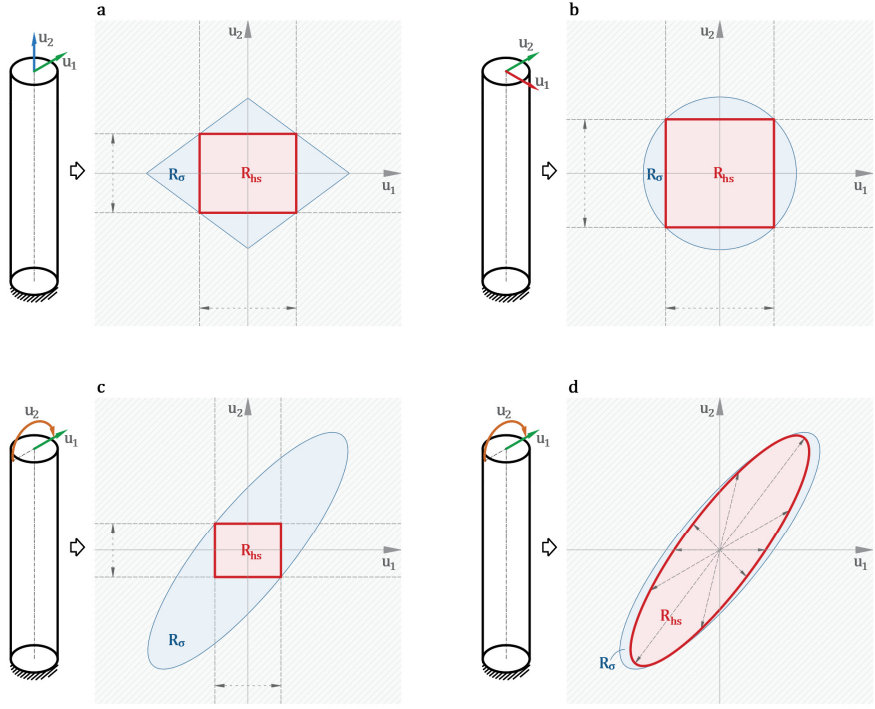

Fig. 1 Examples of two-DOF stress response space $\mathcal{R}_\sigma$ : a) linear-superposition, b) radius-determined, and c, d) synergistic/antagonistic stress response. Corresponding optimal design of hard-stop-free workspace $\mathcal{R}_{hs}$ are shown: a, b, c) multiple single-DOF hard stops, and d) coupled multi-DOF hard stop.

2.3 Hard-stop-free workspace with different DOFs

We next provide a theoretical demonstration of the derivation of the hard-stop-free workspace for different types of hard stop designs: 1) a single-DOF hard stop, 2) multiple independent single-DOF hard stops, and 3) a coupled multi-DOF hard stop. We also present the corresponding shape functions that each design can achieve for the hard-stop-free workspace $\mathcal{R}_{hs}$. It is critical to note that the hard stops must be placed at the same end $A$ where load $W$ is applied, namely, the hard stops must directly constrain the motion of the free-stage end, without any intermediate flexible elements. For example, placing hard stops on each joint of a robotic arm does not directly constrain the motion of the end tool, as the deformation of the intermediate arm segments is not controlled.

A single-DOF hard stop imposes a motion limit on a specific degree of freedom (DOF) within a fixed range. Its corresponding hard-stop-free workspace is defined as

$$\mathcal{R}_{hs}^1(k) = \left\{ \boldsymbol{U}: \boldsymbol{U} = \sum_{i=1}^{K} u_i \boldsymbol{e}_i, u_k \in [a_k, b_k] \right\}, \tag{15}$$

where $k$ denotes the specific DOF to which the motion limit $[a_k, b_k]$ is applied, while all other DOFs remain unconstrained. The shape definition of $\mathcal{R}_{hs}^1(k)$ is given by

$$\mathcal{R}_{hs}^1(k) = [a_k, b_k] \times \prod_{\substack{i=1 \\ i \neq k}}^{K}(-\infty, \infty), \tag{16}$$

which corresponds to a pair of parallel, infinite surfaces in the $K$-dimensional space. Because a single-DOF hard stop constrains motion in only one direction while leaving all others unrestricted, it produces a large "unprotected motion space" $\mathcal{R}_{hs\setminus\sigma}$, thereby increasing the risk of failure under highly uncertain loads.

We next consider a system of multiple single-DOF hard stops, composed of independently placed single-DOF hard stops $\mathcal{R}_{hs}^1(k)$ for each degree of freedom that needs protection, where $k \in \{k_1, k_2, \ldots, k_K\}$. Because these hard stops operate independently, their combination results in a hard-stop-free workspace defined by the intersection of each individual single-DOF workspace:

$$\cap^K \mathcal{R}_{hs}^1 = \bigcap_{k=1}^{K} \mathcal{R}_{hs}^1(k). \tag{17}$$

Explicitly, the shape definition of $\cap^K \mathcal{R}_{hs}^1$ can be written as

$$\cap^K \mathcal{R}_{hs}^1 = \prod_{i=1}^{K} [a_i, b_i], \tag{18}$$

which forms an orthotope in $K$-dimensional space (rectangle in 2-DOF motion, a rectangular cuboid in 3-DOF motion, etc.). Because this hard-stop arrangement does not couple motion limits across different directions, it is inefficient for systems where the stress response involves directional coupling. For example, the two-DOF stress response in Fig. 1a follows a linear superposition and yields a diamond-shaped safe stress space $\mathcal{R}_\sigma$. However, since the 2D shape of $\mathcal{R}_{hs}$ (defined by Eq. (18)) is necessarily rectangular, the maximum achievable volume fraction $\phi_{hs}$ is limited to 0.5, regardless of the specific parameter values used in the design. The limitation of $\cap^K \mathcal{R}_{hs}^1$ worsens when the coupling effect of the stress response is stronger. For example, when the stress response exhibits synergistic and antagonistic interactions among different directional motions (Fig. 1c,), the orthotope shape of $\mathcal{R}_{hs}$ forces the hard-stop design to be overly conservative—limited by the direction in which the mechanism is weakest. This low $\phi_{hs}$ leads to extreme overprotection, characterized by both wearing and mechanical shock during all but a narrow range of loading activities. As such, multiple single-DOF hard stops will only be optimal for systems with orthogonal stress responses (which, as shown, are rare in practice).

To achieve an effective volume fraction of hard-stop-free workspace for systems with *skewed* safe stress spaces, coupled motion limits *across dimensions* are required. Assuming the safe stress space is radially convex, a convenient approach for defining such coupled multi-DOF hard stops is to use *radial constraints* on the multi-DOF motion. In this formulation, instead of specifying motion limits independently along each $e_i$ direction, we define motion constraints along all possible radial directions. First, we define a closed surface $\Gamma^{hs}$ corresponding to the hard-stop engagement—that is, for any $U \in$ $\Gamma^{hs}$, the hard stop makes contact without penetration. The hard-stop-free workspace is then defined as

$$\mathcal{R}_{hs} = \left\{ U : U = k \sum_{i=1}^{K} u_i e_i, 0 < k < 1, u_i e_i \in \Gamma^{hs} \right\}. \tag{19}$$

In this formulation, $\mathcal{R}_{hs}$ is a radially convex volume enclosed by the finite boundary $\Gamma^{hs}$. For example, Fig. 1d shows a multi-DOF hard-stop design for synergistic/antagonistic stress response. Construction of $\mathcal{R}_{hs}$ from radial constraints enables coupled motion limits along arbitrary directions in the mechanism's workspace (i.e. not restricted to $\pm e_i$), paving the way for optimization of $\phi_{hs}$. In this way, multi-DOF hard-stop designs that couple motion limits across different dimensions are positions to be much more effective than multiple single-DOF hard stops for systems with strongly coupled stress responses.

2.4 Shape design of multi-DOF hard stops

The shape design of multi-DOF hard stops that achieve boundary $\Gamma^{hs}$ with coupled limits is inherently more complex than arrangements of multiple single-DOF hard stops, since $\Gamma^{hs}$ generally cannot be determined analytically from the geometry, except in the special case of point-to-surface contact. However, point-to-surface hard stops are not overly practical in real-world designs due to 1) their inability to restrict end rotation, 2) extremely high local contact pressure, and 3) poor post-contact stability. By contrast, surface-to-surface contacts, despite design complexity, can provide both rotational and translational motion limit, minimize peak contact pressure, and improve impact absorption by decomposing contact forces into normal and shear components.

This section presents a numerically implementable framework for designing multi-DOF hard stops via surface-to-surface contact (implementation details are exemplified in case study, Sec. 3.4). First, we define $\Omega_a$ and $\Omega_b$ as a pair of hard-stop surfaces located on the free-stage end $A$ and the ground body $B$, respectively, where both $\Omega_a$ and $\Omega_b$ are assumed to be rigid, open surfaces. The position vector of an arbitrary material point on the stage hard stop $\Omega_a$ is defined as $X_a$ in the undeformed (reference) configuration, and $x_a$ in the deformed configuration, after a rigid body motion of $\Omega_a$ by deformation vector $U$:

$$x_a = \chi(X_a, U), \tag{20}$$

where $\chi$ defines the mapping relationship between the undeformed and deformed configurations of a material point. The condition for contact-free motion after the rigid transformation $U$ of $\Omega_a$ is given by:

$$d(x_a, \Omega_b) > 0, \forall x_a \in \Omega_a, \tag{21}$$

where $d(x_a, \Omega_b)$ represents the shortest distance from point $x_a$ to the surface $\Omega_b$, defined as

$$d(x_a, \Omega_b) = \inf\{\|x_a - x_b\| : x_b \in \Omega_b \}, \tag{22}$$

where $\inf\{\cdot\}$ defined the infimum of the set. Note that here it is assumed that the ground hard stop $\Omega_b$ is sufficiently large relative to $\Omega_a$. Accordingly, the hard-stop-free workspace for the surface-to-surface contact pair is defined as:

$$\mathcal{R}_{hs} = \{U: \forall X_a \in \Omega_a, d(\chi(X_a, U), \Omega_b) > 0\}. \tag{23}$$

To rewrite this in the radial form of motion limits, we first need to obtain the boundary $\Gamma^{hs}$ by surface collision:

$$\Gamma^{hs} = \{U: \forall X_a \in \Omega_a, d(\chi(X_a, U), \Omega_b) = 0\}. \tag{24}$$

Using this framework, for a given pair of hard-stop surfaces $\Omega_a - \Omega_b$, a collision simulation can first be performed to determine the contact boundary $\Gamma^{hs}$. Then, the corresponding hard-stop-free workspace $\mathcal{R}_{hs}$ can be obtained via Eq. (19). Next, based on the derived $\mathcal{R}_{hs}$ and $\mathcal{R}_\sigma$, the "unprotected motion space" $\mathcal{R}_{hs\setminus\sigma}$ (Eq. (9)) and "overprotected motion space" $\mathcal{R}_{\sigma\setminus hs}$ (Eq. (11)) can be obtained to evaluate the performance of the hard-stop shape $\Omega_a - \Omega_b$ according to the criteria in Eq. (10) and Eq. (12).

2.5 Hard-stop design framework

Based on the discussion above, we propose a general hard-stop design framework for compliant mechanisms, consisting of the following steps:

1) Stress response analysis: Determination of $\mathcal{R}_\sigma$ based on the stress response function $\sigma^{max} = \mathbb{R}(U)$ of the compliant mechanism.

2) Hard-stop configuration: Determination of the hard-stop arrangement and contact pair shapes best suited to the derived safe stress space.

3) Parameter optimization: Optimization of the hard-stop shape parameters, based on maximizing the volume fraction $\phi_{hs}$ while ensuring sufficient protection $\mathcal{R}_{hs} \subseteq \mathcal{R}_\sigma$.

In the following sections, we apply this design framework to a representative case study. We demonstrate the full end-to-end design process, including understanding the hard-stop requirements, mapping the mechanism's stress response, generating the hard-stop geometries, and validating the design through numerical simulation and experimental testing.

## 3. Representative Case Study

3.1 Case overview

The representative compliant mechanism is a TKA component (Fig. 2a), which works as an intramedullary stem to connect the central joint-replacing structure to the proximal tibia bone[37]. This mechanism, which we consider to be quasi-circular-symmetric, comprises a 22-blade caged hinge that is inverted via a flanged post, as shown Fig. 2b, such that compression of the knee joint creates tension in the mechanism. This mechanism provides high compliance in rotation about the cylinder's long axis (reaction to tangential loading), moderate compliance in radial (lateral) directions, and low compliance in the axial (vertical) direction. The tangential compliance, which serves as the key function of the design, reduces the shear loads on the bone-implant interface and prevents cross-shearing in the articular bearing during internal/external tibial rotation, thereby increasing the lifespan of the joint replacement.

When the knee is loaded during lower-extremity tasks, the mechanism transfers these loads from the femur to the tibia. The flange top face (Fig. 2c) is regarded as the free-stage end $A$, and we consider it as a rigid body coupled with a load reference point $O_L$. The mechanism is suspended within a rigid case installed in the tibia bone, with the interface between the mechanism and the case treated as a grounded end $G$. Defining the stem's longitudinal axis as the z-axis and the two orthogonal lateral axes as $x$ and $y$, we consider a combined $S = 6$ DOF load vector, $W = (F_x, F_y, F_z, M_x, M_y, M_z)$, applied at the load reference point $O_L$ (Fig. 2c), where the Cartesian $x$-, $y$-, and $z$-axes correspond to the medio-lateral (ML), anterior-posterior (AP), and longitudinal axes, respectively (Fig. 2a). In this case study, we use published knee loading data from instrumented TKA implants for tasks including walking, knee bending, quiet standing, sit-to-stand, stair descent, stair ascent, stand-to-sit, and jogging based on average load levels for a subject weighing $75\ kg$ [51–57].

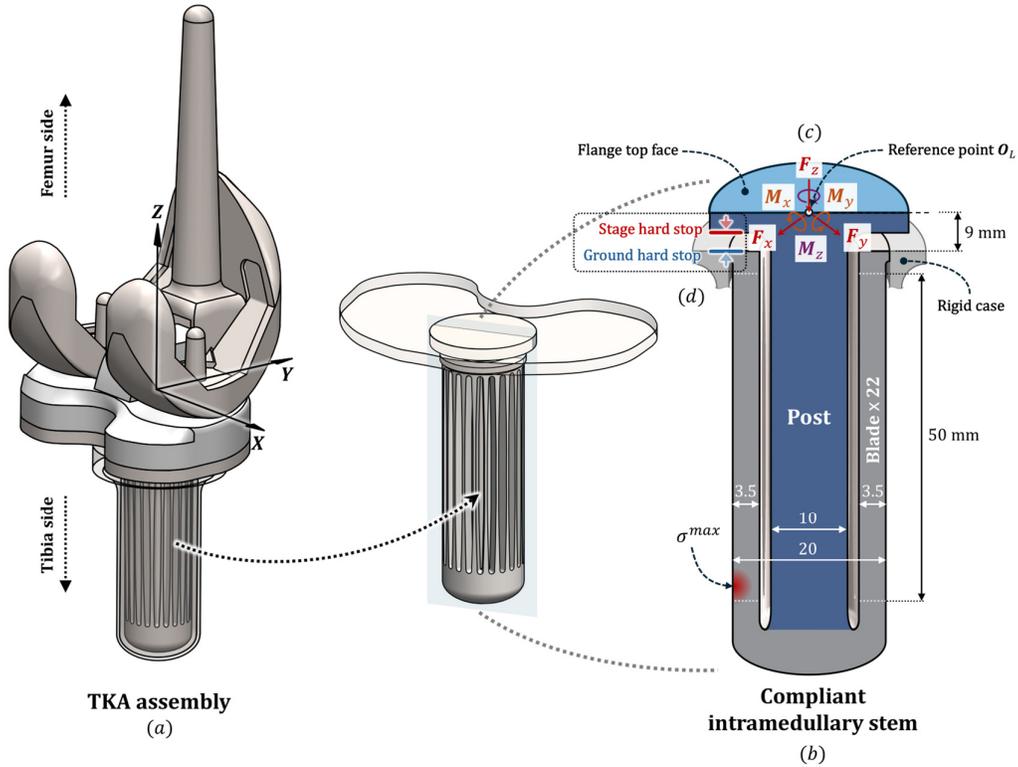

Fig. 2 a) Total knee arthroplasty (TKA) implant with a compliant intramedullary stem on the tibial side. b) Compliant intramedullary stem with invertedly connected caged hinge and flanged post.

3.2 Determination of hard-stop workspace

The free-stage hard-stop surface $\Omega_a$ is constructed on the load input flange as a rigid shell, while the ground-side hard-stop surface $\Omega_b$ is built on top of the rigid case (Fig. 2d). We consider the motion of the reference point $O_L$ as $U_S = (\delta_x, \delta_y, \delta_z, \vartheta_x, \vartheta_y, \vartheta_z)$, representing the motion with all possible

DOFs in the same Cartesian coordinate system that defines $W$. Here, $\boldsymbol{\delta} = (\delta_x, \delta_y, \delta_z)$ denotes the translational displacement vector, and $\boldsymbol{\vartheta} = (\vartheta_x, \vartheta_y, \vartheta_z)$ is the axis-angle rotational vector. To determine $K$, the total number of DOFs that require protection, we first exclude the vertical translation $\delta_z$ due to sufficient rigidity and strength in the longitudinal direction. Furthermore, based on published datasets, the input load does not prescribe an axial torque component $T = M_z$, but instead reports a range of axial rotation $\vartheta_z$. Therefore, we also exclude hard-stop protection for $\vartheta_z$, and instead fix this displacement at its worst-case value of 7° (clockwise, from a top view) during stress analysis. Note that we do not claim that the mechanism could not possibly be overloaded in these DOFs; rather, we have concluded that failure from overloading in these DOFs is substantially less likely than in the DOFs that we have deemed merit protection. As such, the global DOFs that are included in the hard-stop-protection design are selected as $\delta_x, \delta_y, \vartheta_x$, and $\vartheta_y$ for the representative mechanism.

Next, we will determine the workspace of the hard stop associated with a special coordinate system. Given the rotational transformation is defined by the axis-angle rotation vector $\boldsymbol{\vartheta}$ about the anchor $\boldsymbol{O}_a$. We can write Eq.(20), the general mapping of rigid body motion of $\boldsymbol{\Omega}_a$, as

$$\boldsymbol{x} = \boldsymbol{R}(\boldsymbol{\vartheta})(\boldsymbol{X} - \boldsymbol{O}_a) + \boldsymbol{O}_a + \boldsymbol{\delta}. \tag{25}$$

For a circular symmetric structure, axial rotation $\vartheta_z$ does not affect the hard-stop contact. Thus, we extract the tilt rotation vector $\boldsymbol{\vartheta}_{xy} = (\vartheta_x, \vartheta_y, 0)$ for hard-stop engagement analysis. Similarly, the translation vector is decomposed as $\boldsymbol{\delta} = \boldsymbol{\delta}_{xy} + \boldsymbol{\delta}_z$, where $\boldsymbol{\delta}_{xy} = (\delta_x, \delta_y, 0)$ represents the in-plane *translational* deflection and $\boldsymbol{\delta}_z = (0, 0, \delta_z)$ corresponds to the vertical settlement. Consequently, Eq. (25) can be rewritten as

$$\boldsymbol{x} = \boldsymbol{R}(\boldsymbol{\vartheta}_{xy})(\boldsymbol{X} - \boldsymbol{O}_a) + \boldsymbol{O}_a + \boldsymbol{\delta}_{xy} + \boldsymbol{\delta}_z, \tag{26}$$

where the rotation tensor $\boldsymbol{R}(\boldsymbol{\vartheta}_{xy})$ is given by

$$\boldsymbol{R}(\boldsymbol{\vartheta}_{xy}) = \boldsymbol{I} + \frac{\sin \vartheta_a}{\vartheta_a}[\boldsymbol{\vartheta}_{xy}]_\times + \left(\frac{1 - \cos \vartheta_a}{\vartheta_a^2}\right)[\boldsymbol{\vartheta}_{xy}]_\times^2,$$

where $\vartheta_a$ is the magnitude of the tilt rotation, $\boldsymbol{I}$ is the identity matrix, $[\boldsymbol{\vartheta}_{xy}]_\times$ is the skew-symmetric matrix of $\boldsymbol{\vartheta}_{xy}$, defined by

$$[\boldsymbol{\vartheta}_{xy}]_\times = \begin{bmatrix} 0 & 0 & \vartheta_y \\ 0 & 0 & -\vartheta_x \\ -\vartheta_y & \vartheta_x & 0 \end{bmatrix}.$$

For convenience, we introduce the "*angular* deflection" vector $\boldsymbol{\vartheta}_{xy}^\perp = (-\vartheta_y, \vartheta_x, 0)$, which represents a right-hand rotation of vector $\boldsymbol{\vartheta}_{xy}$ by 90° in the $xy$-plane. This auxiliary vector proves convenient for a circular symmetric mechanism because, under a pure lateral shear force or bending moment, $\boldsymbol{\vartheta}_{xy}^\perp$ aligns with $\boldsymbol{\delta}_{xy}$. The magnitude of the angular deflection is identical to that of the tilt rotation, obtained by

$$\vartheta_a = \|\boldsymbol{\vartheta}_{xy}\| = \|\boldsymbol{\vartheta}_{xy}^\perp\|. \tag{27}$$

Similarly, the magnitude of translational deflection is

$$\delta_a = \|\boldsymbol{\delta}_{xy}\|. \tag{28}$$

Since the hard-stop engagement establishes radially in a non-preferential direction, it is independent of the absolute orientations of $\boldsymbol{\delta}_{xy}$ and $\boldsymbol{\vartheta}_{xy}^{\perp}$ but rather depends on their relative orientations. Therefore, we define a separation angle between the directions of *angular* and *translational* deflection vectors via

$$\theta_{sep} = \cos^{-1} \frac{\boldsymbol{\delta}_{xy} \boldsymbol{\vartheta}_{xy}^{\perp}}{\|\boldsymbol{\delta}_{xy}\| \|\boldsymbol{\vartheta}_{xy}^{\perp}\|}, \tag{29}$$

when $\vartheta_a \delta_a \neq 0$. Furthermore, we consider the vertical settlement of the hard stop $\delta_z$ as a reference parameter determined by different levels of vertical compressive force $F_z$. Finally, the workspace of the hard stop can be written as

$$\boldsymbol{U} = (\delta_a, \vartheta_a, \theta_{sep}) \tag{30}$$

in a special three-DOF space, defined by axes: 1) translational deflection $\delta_a$, 2) angular deflection $\vartheta_a$ and 3) separation angle $\theta_{sep}$.

By slicing the 3D workspace of along the $\delta_a - \vartheta_a$ plane at a constant $\theta_{sep}$, we obtain a 2D representation of the workspace. For example, the 2D workspace shown in Fig. 3b plots the workspace of the hard stop at a fixed separation angle $\theta_{sep}$, where the horizontal and vertical axes represent the amplitude of the translational deflection $\delta_a$ and angular deflection $\vartheta_a$. By convention in this manuscript, the first and third quadrants correspond to the acute separation angle $\theta_{sep} \in [0°, 90°]$, while the second and fourth quadrants correspond to the supplementary angles $(180° - \theta_{sep})$. Therefore, the workspaces of hard stop presented in this paper are shown with $\theta_{sep}$ and $(180° - \theta_{sep})$ in the same plot. Each point in Fig. 3b corresponds to a specific motion vector $\boldsymbol{U} = (\delta_a, \vartheta_a, \theta_{sep})$ of the stage hard stop. For instance, the motion shown in Fig. 3a corresponds to the black point in Fig. 3b, defined by translational deflection $\delta_a$ and angular deflection $\vartheta_a$ at the separation angle $\theta_{sep}$. The red area shown in Fig. 3b represents the hard-stop-free workspace $\mathcal{R}_{hs}$, and its boundary $\boldsymbol{\Gamma}^{hs}$ is plotted as the red stroke lines. The characteristic parameters $\vartheta_{max}$, $\Delta\vartheta_\delta$ and $\delta_{max}$ (Fig. 3b) define the *shape* of the hard-stop-free workspace, which is primarily determined by the shape function of the hard-stop surfaces.

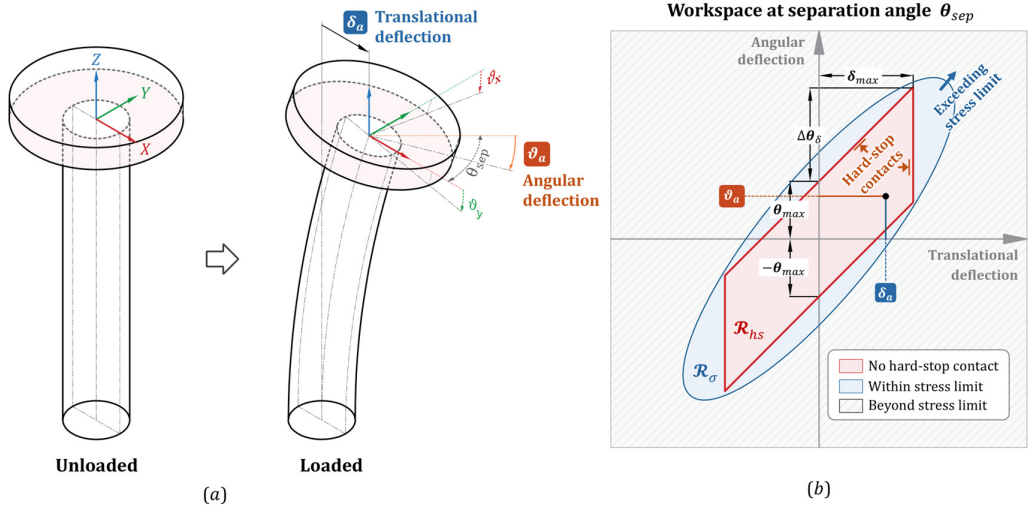

Fig. 3 a) Illustration of motion of the stage hard stop $\Omega_a$ with angular deflection $\vartheta_a$ and translational deflection $\delta_a$ at separation angle $\theta_{sep}$.

b) Illustration of hard-stop-free workspace $\mathcal{R}_{hs}$ and safe stress space $\mathcal{R}_\sigma$ in the $\delta_a - \vartheta_a$ plane at separation angle $\theta_{sep}$.

3.3  Analysis of stress response function

Given that our design objective is to achieve a fatigue life of $10^8$ cycles (approximately 100-year life), we define the stress response as the maximum *principal* stress $\sigma^{max}$ within the mechanism, and seek to keep this stress below the material fatigue limit. Based on Eq. (30) and Eq. (4), the stress response can be expressed as:

$$\sigma^{max} = \mathbb{R}(\boldsymbol{U}) \\ = \mathbb{R}(\delta_a, \vartheta_a, \theta_{sep}), \tag{31}$$

where we consider the vertical compressive force $F_z$ given at a fixed value. Based on Eq.(31) we can derive the safe stress space $\mathcal{R}_\sigma$ (Eq. (7)) enclosed by $\boldsymbol{\Gamma}^\sigma$ Eq. (6), within which any motion $\boldsymbol{U}$ of stage hard stop does not result in the mechanism exceeding the stress limit $\sigma^{cr}$.

Given that the stress response function in Eq. (31) is both highly nonlinear and implicit, we choose to define the function using numerical methods such as finite element analysis (FEA). In this section, we implement the finite element model in ABAQUS (ABAQUS 2023, Dassault Systems, Velizy-Villacoublay, France) to simulate stress responses for different vertical compressive forces $F_z$, expressed as $\sigma^{max}|_{F_z} = \mathbb{R}(\delta_a, \vartheta_a, \theta_{sep})$. Three different levels of $F_z$ are considered, corresponding to normal loading, moderate overloading, and high overloading cases. The normal loading is set at $F_z = 3kN$ based on the rounded upper limit of the maximum vertical compressive force ($F_z^{max} = 2937.6N$) observed across six typical knee activities [51–57]. The moderate and high overloading cases are considered as $2\times$ and $3.3\times$ amplifications of normal loading, yielding $F_z = 6kN$ and $F_z = 10kN$. For all considered cases, the axial torque $T$, which will be constrained by a simple rotational hard stop in the final implant [37], is preloaded at its worst state (based on the same knee activity data) corresponding to a fixed torsion of $\vartheta_z = -7°$. The stage hard stop, along with the flanged post and caged hinge, is modeled using Ti-6Al-4V (Grade 5) titanium alloy, while the ground hard stop is modeled with GUR 1020 UHMW-PE polyethylene [58]. Additional details of the FEA modeling can be found in Appendix. B. For the stress response analysis in this section, we deactivate the contact pairs in the FEA simulations, allowing the flanged post to move freely according to the specified motion vector.

Fig. 4 presents the results of maximum principal stress in the mechanism as heatmaps, which are plotted in the 2D workspace of the hard stop ($\delta_a - \vartheta_a$ plane) under given motion vectors and vertical compressive forces. In this study, we adopt a conservative fatigue criterion, corresponding to the Ti-6Al-4V $10^8$-cycles fatigue limit of $480 MPa$ [39,40], which theoretically guarantees a 100-year life. The contour corresponding to the fatigue limit ($\sigma^{max} = 480 MPa$) is plotted as continuous red curves, and the enclosed area is the safe stress space for fatigue $\mathcal{R}_\sigma|_{fatigue}$. In contrast, the contour corresponding to the yielding limit ($\sigma^{max} = 880 MPa$) are plotted as dashed red curves, and the enclosed area is defined as the safe stress space for yielding $\mathcal{R}_\sigma|_{yield}$. It is important to note that while the maximum principal stress is typically used for fatigue estimation, von Mises stress is more appropriate for analyzing the yield limit. For unified illustration, the yielding limit is plotted using the maximum principal stress in Fig. 4. However, the actual $\mathcal{R}_\sigma|_{yield}$ derived and used for hard-stop design in this study is based on von Mises stress.

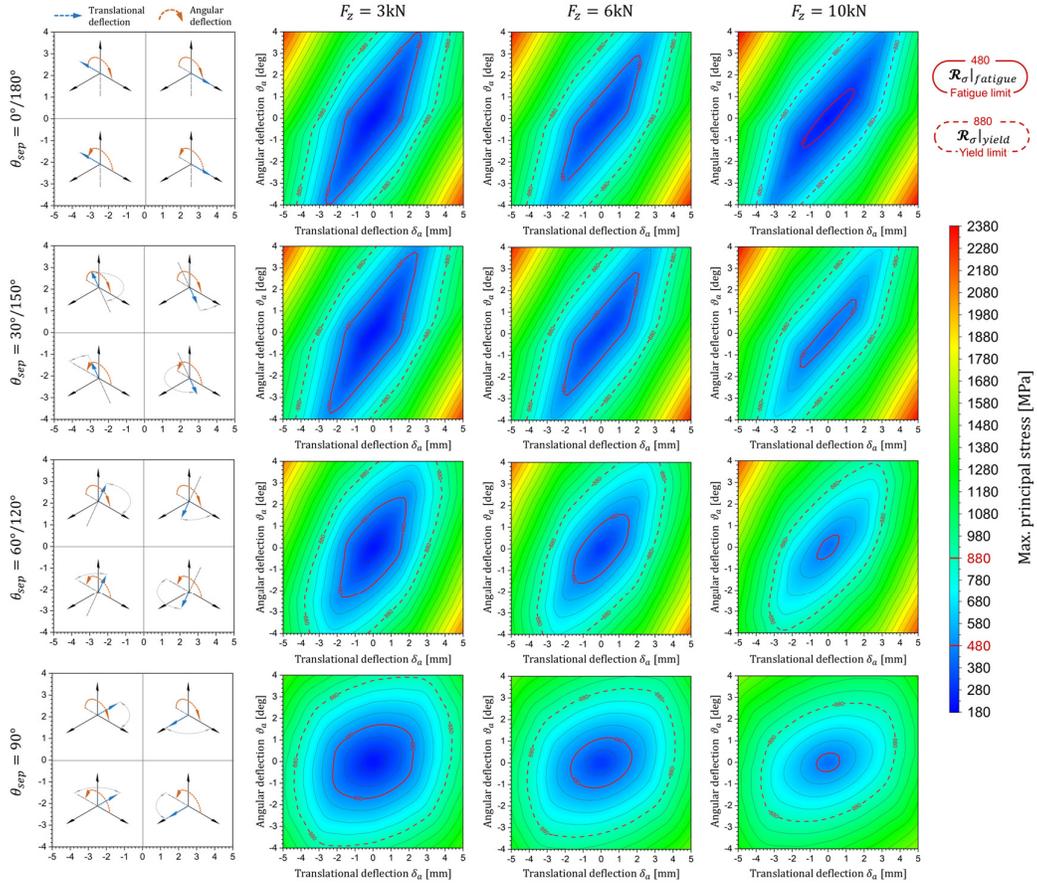

Fig. 4 Heatmaps of maximum principal stress in the compliant intramedullary stem as a function of translational ($\delta_a$) and angular ($\vartheta_a$) deflections, shown in 4 × 3 grid layout: horizontal grids for compressive forces $F_z = 3kN$, $6kN$ and $10kN$ (from left to right); vertical grids for separation angles $0°/180°$, $30°/150°$, $60°/120°$, and $90°$ (from top to bottom). Each heatmap's quadrants indicate positive/negative amplitudes of $\delta_a$ and $\vartheta_a$: first and third quadrants correspond to acute separation angles; second and fourth quadrants correspond to the supplementary angles.

From Fig. 4, we observe that $\mathcal{R}_\sigma$ in the $\delta_a - \vartheta_a$ planes exhibits a skewed shape across nearly all values of $F_z$ and $\theta_{sep}$. The stress increases more rapidly with the motion vectors in the second and fourth quadrants, much faster than that in the first and third quadrants. This "skewed" $\mathcal{R}_\sigma$ arises because the effects of translational and angular deflections on maximum stress can either be synergistic or antagonistic, depending on the separation angle $\theta_{sep}$. When $\theta_{sep} < 90°$, the translational and angular deflections have an acute separation angle, their antagonistic effect on stress becomes dominant, which relaxes the stress increase. Conversely, when the translational and angular deflections have an obtuse separation angle $\theta_{sep} > 90°$, as in the second and fourth quadrants, their synergistic effect rapidly increases the stress beyond the allowable stress limit. Interestingly, the degree of skewness decreases as $\theta_{sep}$ varies following the sequence $0°/180°$, $30°/150°$, $60°/120°$, and $90°$. However, a slightly skewed $\mathcal{R}_\sigma$ is still evident in the last row of heatmaps in Fig. 4, corresponding to perpendicular translational and angular deflections—where their effects should be symmetric. This can be attributed to the mechanism being subjected to a loaded torsion $\vartheta_z = -7°$, which breaks the initial symmetry of the blade distribution. In particular, in our FEA results, when the axial rotation is fixed at $\vartheta_z = 0°$, a symmetric $\mathcal{R}_\sigma$ is observed for $\theta_{sep} = 90°$. For the same $\theta_{sep}$, the degree of skewness does not change significantly with an increase in $F_z$; instead, increasing $F_z$ leads to a gross reduction in the size of $\mathcal{R}_\sigma$. This stems from the fact that higher $F_z$ generates higher base normal stress within the blades, which is later redistributed by the angular and translational deflections.

3.4 Hard-stop shape design

We seek to design hard-stop shapes that effectively maximize $\mathcal{R}_{hs}$ within the skewed space $\mathcal{R}_\sigma$ (Fig. 4); this involves multi-DOF hard-stop design that minimizes wear while ensuring effective stress protection ($\mathcal{R}_{hs} \subseteq \mathcal{R}_\sigma$). The shape design can be employed using multiple approaches, including explicit parametric optimization and implicit topology optimization. In this study, we adopt a parametric optimization approach, wherein the shape function of $\Omega_a$ and $\Omega_b$ are predefined and their parameters are optimized to approximate the solution to Eq. (12) and Eq. (9). As demonstrated in Sec.2.3, the maximum achievable volume fraction $\max(\phi_{hs})$ depends on the *shape functions* of both the $\mathcal{R}_{hs}$ and $\mathcal{R}_\sigma$. If $\mathcal{R}_{hs}$ and $\mathcal{R}_\sigma$ share a similar shape but differ only in dimensional parameters, achieving $\max(\phi_{hs})$ close to 1 becomes straightforward. Given the skewed nature of $\mathcal{R}_\sigma$, our hard-stop design should select $\Omega_a$ and $\Omega_b$ geometries that can provide a similarly skewed hard-stop-free workspace $\mathcal{R}_{hs}$. For a circular symmetric shape, we consider three possible hard-stop shape functions: 1) "simple shaft-hole", 2) "flanged shaft-hole", and 3) "torus caps", as illustrated in the space of $(\delta_a, \vartheta_a)|_{F_z, \theta_{sep}=0°/180°}$ (Fig. 5).

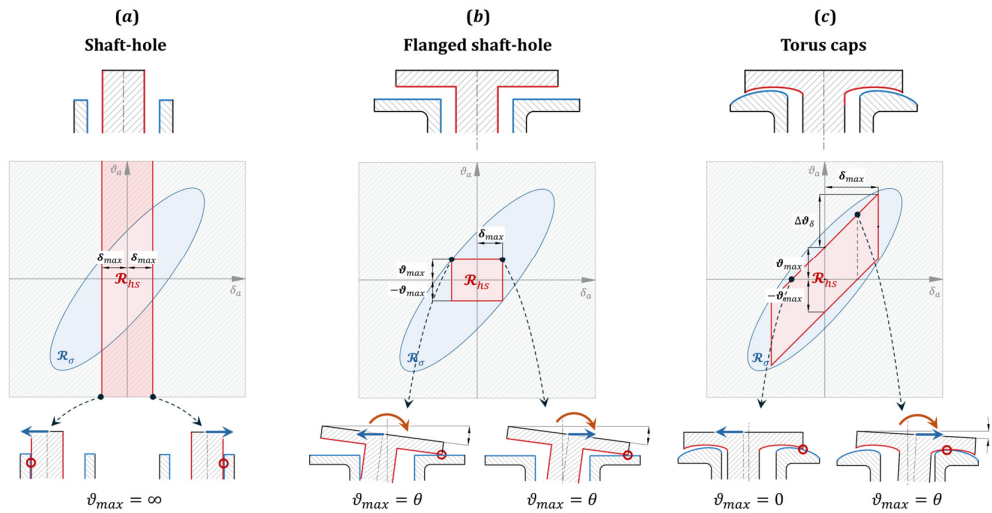

Fig. 5 Hard-stop shapes and their corresponding hard-stop-free workspaces: a) Simple shaft-hole; b) Flanged shaft-hole; c) Torus caps.

The simple shaft-hole hard stop (Fig. 5a) functions as a single-DOF hard stop, only capable of constraining translational deflection while leaving angular deflection unrestricted. This shape design is suitable only for mechanisms that are not sensitive to bending moments. The flanged shaft-hole hard stop (Fig. 5b) *independently* constrains translational and angular deflections. This configuration can be regarded as an arrangement of multiple single-DOF hard stops, which provides only a rectangular $\mathcal{R}_{hs}$ with a very limited volume fraction $\phi_{hs}$, leading to unnecessary wearing of the hard stop. As discussed in Sec. 2.3, overcoming this limitation requires coupled, multi-DOF hard stops. Following the general design principles for multi-DOF hard stops discussed in Sec. 2.4, we propose a circularly symmetric hard-stop configuration, based on so-called "torus caps." This shape function imposes constraints in the radial direction $(\delta_a, \vartheta_a)$ by coupling critical angular deflection to the magnitude of translational deflection. For example, if the stage hard stop experiences a large enough translational deflection to the left (Fig. 5c), the angular deflection to the right is restricted to zero. Conversely, if the stage hard stop moves right, the angular deflection in the same direction thus gains a larger space to accommodate. In effect, the maximum allowable angular deflection increases with the translational deflection in the same direction. As a result, we find that the torus cap configuration produces a skewed $\mathcal{R}_{hs}$ that spreads more into the first and third quadrants (see later results in Fig. 8). Moreover, the torus cap hard-stop surface decomposes impact forces into tangential sliding - similar to a ricochet effect - enhancing shock absorption while maintaining internal stress levels within safe limits. Thus, a well-designed "torus cap" hard stop is considered ideal for maximizing $\phi_{hs}$ in a skewed safe stress space $\mathcal{R}_\sigma$.

*Oblique elliptic torus hard stop*

The oblique elliptic torus hard stop is a specialized shape function classified as a "torus cap", formed by revolving an ellipse with an oblique axis. This shape function also offers large optimization flexibility without an overwhelming number of parameters, making it the preferred choice for this case study. To formalize the hard-stop shape design, we represent the torus caps as a common initial shape function for both the stage and ground hard stops, beginning with the undeformed (reference) configuration. The Cartesian position vector $\boldsymbol{X}_a = \{X(u,v), Y(u,v), Z(u,v)\}$ defines an arbitrary material point on hard-stop surface $\boldsymbol{\Omega}_a$

$$\begin{aligned} X(u,v) &= X'(u)cos(v), \\ Y(u,v) &= X'(u)sin(v), \\ Z(u,v) &= Z'(u). \end{aligned} \quad (32)$$

Eq. (32) describes a torus cap, formed by revolving the upper portion of an oblique ellipse in the $x-z$ plane around the $z$-axis. The ellipse curve is defined by:

$$X'(u) = (u - R_C)\cos(\theta_O) - \frac{d_S}{d_L}\tan(\theta_O)\sqrt{\frac{d_L^2}{4} - (u - R_C)^2} + R_C,$$

$$Z'(u) = (u - R_C)\sin(\theta_O) + \frac{d_S}{d_L}\sqrt{\frac{d_L^2}{4} - (u - R_C)^2},$$

(33)

where $u$ is given in range $u \in [R_C - d_L/2, R_C + d_L/2]$, as shown in Fig. 6a. Here: $d_L^{(i)}$ and $d_S^{(i)}$ denote the ellipse's major and minor axes, respectively; $R_C^{(i)}$ represents the radial distance from the ellipse center to the central z-axis; $\theta_O^{(i)}$ is the oblique inclination angle of the ellipse's major axis, defined as a counterclockwise rotation in $x - z$ plane. The superscript $(i)$ indicates the hard-stop type: $(i) = a$ refers to the stage hard stop, and $(i) = b$ refers to the ground hard stop. $z_{ab}$ represents the vertical distance between the ellipse centers of stage and ground hard stops. Moreover, to avoid a sharp contact between the stage hard stop's outer edge and the ground hard-stop surface, we clip the stage hard stop $\Omega_a$ to a specified diameter $d_{hs}^a$ (Fig. 6a).

For the stage hard stop $\Omega_a$, a rigid transformation $U = (\delta_a, \vartheta_a, \theta_{sep})$ is applied. The anchor of the rotational transformation, $O_a$, is aligned with the cut surface of the proximal tibia bone and positioned on the central axis at the same height as the topmost point of the stage hard stop (Fig. 6b). Consequently, $O_a$ has an offset height $z_{oa}$ above the ellipse center of the stage hard stop. The load reference point $O_L$ is defined $z_{Lo} = 9mm$ above $O_a$, accounting for the thickness of the titanium stage hard stop ($3\ mm$) and the typical thickness ($6\ mm$) of the tibial polyethylene insert [59]. Finally, the transformed position vector $x = \{x(u, v), y(u, v), z(u, v)\}$ of an arbitrary material point on $\Omega_a$ in the deformed configuration is given as:

$$x(u, v) = [X'(u)\cos(v)\cos(\vartheta_a) + (Z'(u) - z_o^a)\sin(\vartheta_a)] + \delta_a \cos(\theta_{sep}),$$
$$y(u, v) = X'(u)\sin(v) + \delta_a \sin(\theta_{sep}),$$
$$z(u, v) = [-X'(u)\cos(v)\sin(\vartheta_a) + (Z'(u) - z_{oa})\cos(\vartheta_a)] + z_{oa}.$$

(34)

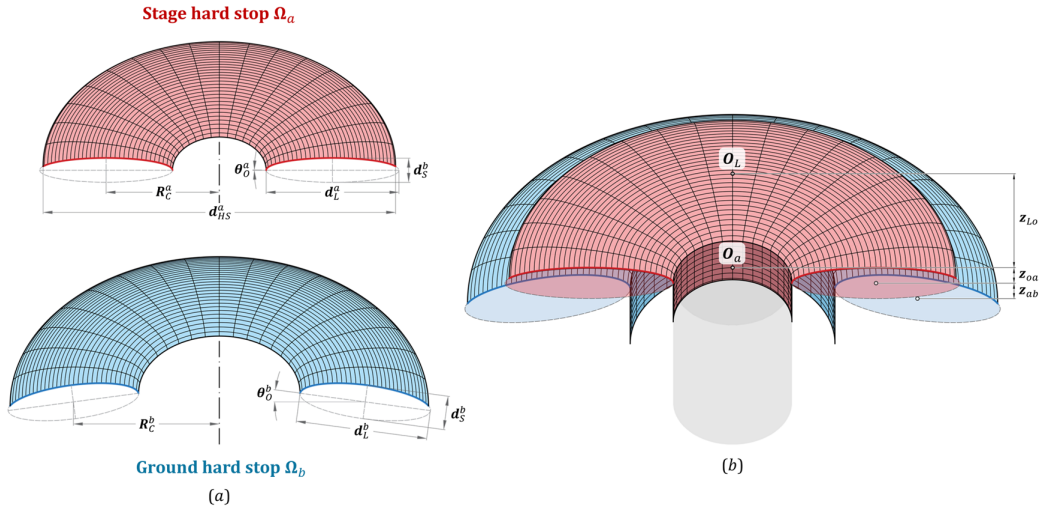

Fig. 6 a) Oblique elliptic torus shapes for the stage (upper) and ground (lower) hard stops.

b) Schematic representation of the hard-stop surface pair in the compliant intramedullary stem.

Next, we solve for the boundary $\Gamma^{hs}$ of the hard-stop-free workspace $\mathcal{R}_{hs}$ defined by Eq. (24). The solution of Eq. (24) on the oblique elliptic torus hard stop is a complex implicit equation that is challenging to derive analytically. In this section, we employ a numerical approach based on particle collision simulations implemented in Python 3.10 to compute the hard-stop-free workspace $\mathcal{R}_{hs}$. The stage and ground hard stops are modeled as finite sets of particles, represented by red and blue particles in Fig. 7, respectively. The particle set of the stage hard stop is applied to the transformation $\boldsymbol{U}$ with all particles coupled to the anchor point $\boldsymbol{O}_a$. The boundary of the hard-stop-free workspace, $h(\boldsymbol{U}) = 0$, is determined using collision detection between the stage and ground hard stops. For clarity, the particle density shown in Fig. 7 is reduced for illustrative purposes. The actual particle density exceeds 64 particles per unit area (1 $mm^2$), with at least $5 \cdot 10^4$ particles on each surface, to ensure accurate computation.

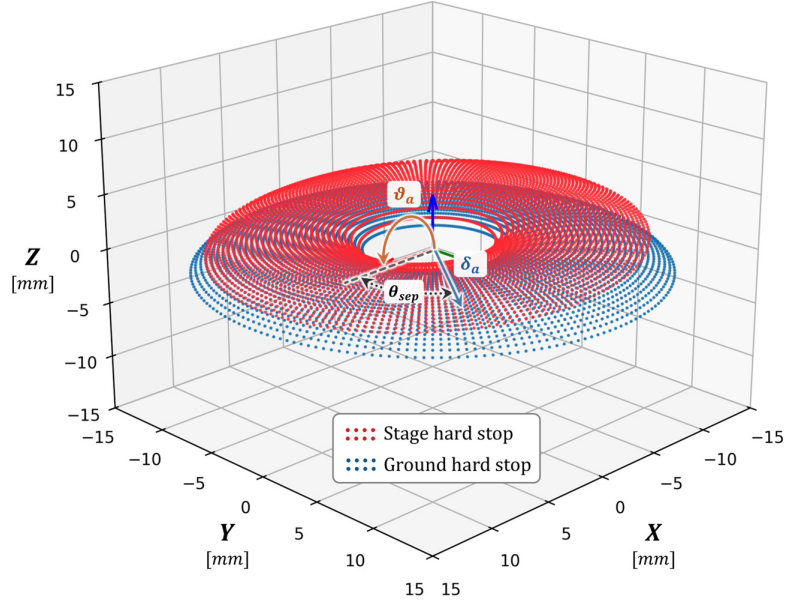

Fig. 7 Simplified illustration of the particle collision simulation for determining the hard-stop-free workspace.

We proceed with the numerical computation to identify the optimal parameters for the oblique elliptic torus hard stop, using the collision simulation approach described above. The objective is to obtain a hard-stop shape that provides a sufficiently large volume fraction $\phi_{hs}$ while ensuring that the safe stress space remains within $\mathcal{R}_\sigma|_{fatigue}$ under normal and moderate-overload conditions. In our computational algorithm, the primary constraint objective $\mathcal{R}_{hs} \subseteq \mathcal{R}_\sigma$ (Eq.(8)) is not enforced as a strict upper boundary, but is instead implemented through penalty functions in each radial direction. Inherently, this penalty method cannot perfectly guarantee $\mathcal{R}_{hs} \subseteq \mathcal{R}_\sigma$ and eliminate the unprotected motion space $\mathcal{R}_{hs\setminus\sigma}$. However, given that the volume of potential $\mathcal{R}_{hs\setminus\sigma}$ is sufficiently small and corresponds to low-frequency overload motions, the impact of numerical imperfection on fatigue life is expected to be negligible. Additionally, we considered another strict constraint: ensuring zero hard-stop engagement along normal-load activity trajectories. Finally, the optimal results are given in Table 1.

Table 1 Geometric parameters of the optimal hard-stop design I

| Stage hard stop | Ground hard stop | Vertical relations |
|---|---|---|
| $d_L^a = 11.4$ | $d_L^a = 11.4$ | $z_{ab} = 0.6645$ |
| $d_S^a = 4.0$ | $d_S^a = 4.0$ | $z_{oa} = 2.0$ |
| $R_C^a = 10.18$ | $R_C^a = 12.129$ | |
| $\theta_O^a = -0.2°$ | $\theta_O^a = -9°$ | |
| $d_{hs}^a = 29.1$ | | |

The results of Design I are presented in three plots in Fig. 8a, b and c, corresponding to different compressive force levels: $F_z = 3kN$ (maximum normal load), $6kN$ (moderate overload), and $10kN$ (high overload), respectively. Note that, we only show the results for $\theta_{sep}$ ranging from $0°$ to $180°$, as the results are periodic with a period of $180°$. Additionally, in Fig. 8a, corresponding to the normal load case, the motion trajectories of various normal activities—such as walking, sit-to-stand, stair ascent, stair descent, stand-to-sit, and jogging—are plotted over a cycle in terms of $\delta_a$, $\vartheta_a$ and $\theta_{sep}$.

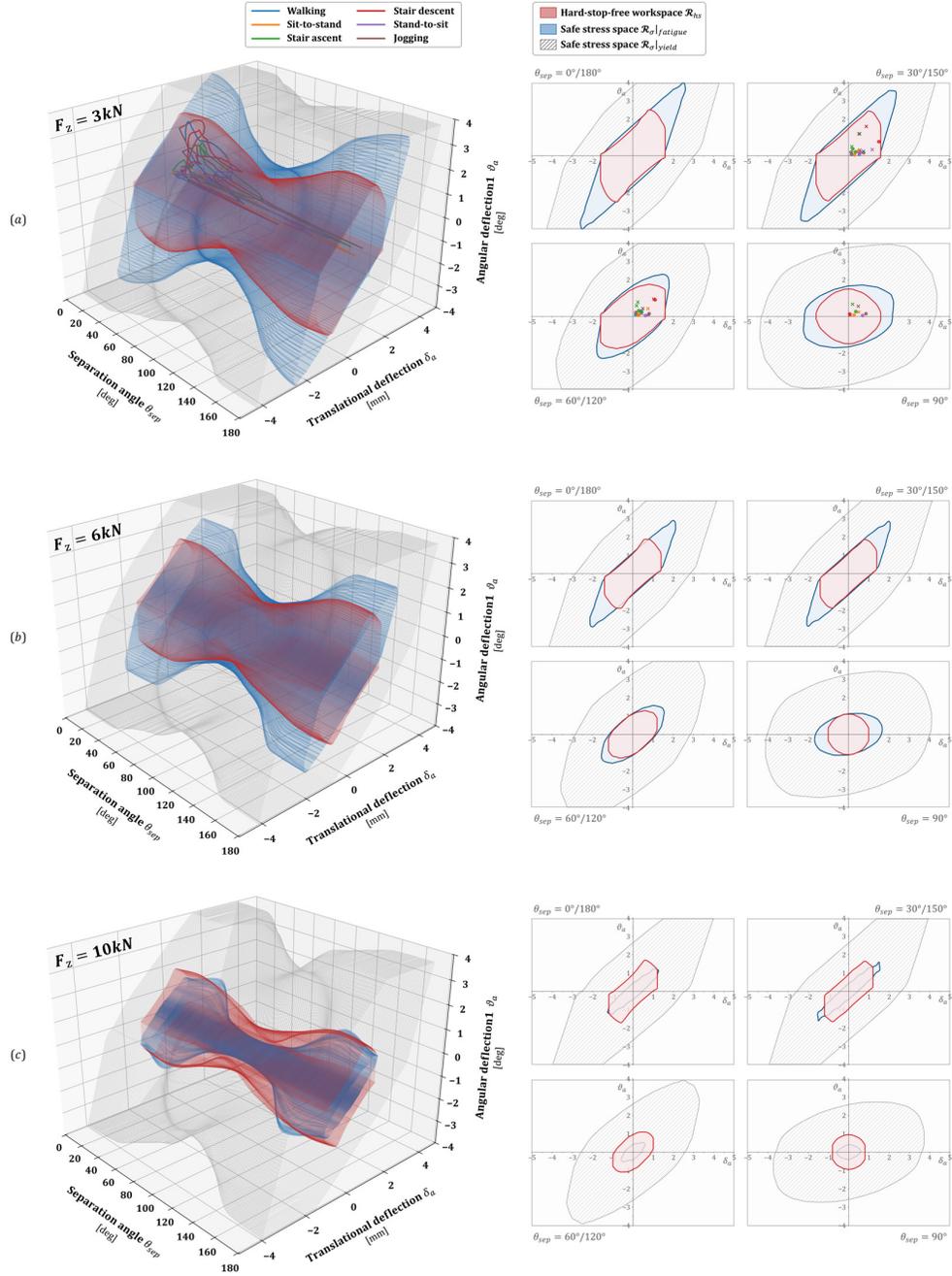

Fig. 8 Hard-stop design results under compressive force levels a) $F_z = 3kN$ (maximum normal load), b) $6kN$ (moderate overload), and c) $10kN$ (high overload).

Left column plots: hard-stop design visualized in 3-DOF workspace defined by $\theta_{sep}$ (left horizontal axis), $\delta_a$ (right horizontal axis), and $\vartheta_a$ (vertical axis).

Right column plots: $\delta_a - \vartheta_a$ cross-sectional planes at specific separation angles $\theta_{sep} = 0°/180°$, $30°/150°$, $60°/120°$, and $90°$.

Red surface: hard-stop-free workspace ($\mathcal{R}_{hs}$). Blue surface: safe stress space for fatigue limit ($\mathcal{R}_\sigma|_{fatigue}$). Grey surface: safe stress space for yield limit ($\mathcal{R}_\sigma|_{yield}$).

As observed in Fig. 8a, design I ensures that $\mathcal{R}_{hs}$ includes all the motion trajectory curves, guaranteeing free motion without any contact or rubbing during all normal activities with different frequencies. Note that $\mathcal{R}_{hs}$ (Fig. 8a) is conservatively based on worst cases in normal activities: $F_z = 3kN$ and $\vartheta_z = -7°$, which is higher than most real-world normal activity loads. The design also achieves a hard-stop-free workspace $\mathcal{R}_{hs}$ that works almost entirely within $\mathcal{R}_\sigma|_{fatigue}$ under $F_z = 3kN$ (Fig. 8a) and $F_z = 6kN$ (Fig. 8b). This ensures that, for overloads up to $2\times$ the normal level, the maximum principal stress in the mechanism remains within the fatigue limit. Moreover, the volume fraction is also sufficiently maximized to $\phi_{hs} = 0.701$, and $\phi_{hs} = 0.747$ under $F_z = 3kN$ and $F_z = 6kN$, respectively, minimizing the wear of the hard-stop surface for overloads that do not surpass the fatigue limit.

However, under high overload condition $F_z = 10kN$ (Fig. 8c), the current design cannot guarantee $\mathcal{R}_{hs}$ is sufficiently contained within $\mathcal{R}_\sigma|_{fatigue}$. This indicates that, under a compressive overload of more than $3\times$, there are loading situations in which the mechanism may experience stresses beyond the fatigue limit without hard-stop engagement. That said, Fig. 8c shows that, $\mathcal{R}_{hs}$ is larger than $\mathcal{R}_\sigma|_{fatigue}$ and remains well within $\mathcal{R}_\sigma|_{yield}$ along with a high safety factor. This implies that, even under very high overloads, design I ensures that the maximum von Mises stress in the mechanism does not surpass the yield limit. Since overloads exceeding $3\times$ the normal level is considered rare, low-frequency incidents that will not significantly impact the mechanism's fatigue life, we have opted not to pursue further optimization to tighten $\mathcal{R}_{hs}$, as this would introduce additional wear to the hard-stop surface under other high-frequency activities.

In summary, the results presented in this section demonstrate that the hard-stop design shown in Table 1 effectively 1) guarantees that the working stress from normal to moderate overloads remains under the material fatigue limit, 2) prevents material yielding under high overloads, and 3) avoids hard-stop engagement during normal activities, and 4) minimizes hard-stop engagement frequency under overloads that do not surpass stress limit. It is important to note that, due to the complexity of the problem, multiple optimal designs may exist with similar performance in engineering practice. The results shown in Table 1 represents one such optimal design.

Nevertheless, the design results presented above are theoretical evaluations of hard-stop performance. However, in real-world applications, additional factors may influence the behavior of the hard stop, which cannot be perfectly resolved in the analytical design framework. Among these factors, the 1) kinematic behavior after hard-stop engagement and 2) extra compliance in the system are of primary interest. With regard to the first of these factors, we assume that the stage hard-stop motion will not escape the hard-stop-free workspace $\mathcal{R}_{hs}$ but may "slide" along its boundary, absorbing impact without the mechanism exceeding the allowable stress limit. Expressed visually, if a motion curve reaches the boundary of $\mathcal{R}_{hs}$ (the red surface in Fig. 8a), it will "adhere" to the surface and continue to move along it without crossing the boundary. However, this theoretical assumption relies on rigid contact, which requires further validation. The second factor involves un-modeled compliance in the system, including the deformation of any connecting pieces or the polyethylene component of the hard stop itself. Therefore, in this section, we conduct numerical and experimental validations to evaluate how real-world factors may cause deviations from analytical predictions and affect actual performance.

3.5  Numerical validation

In this section, we use FEA (ABAQUS 2023) to validate the hard-stop performance under a short-period overload surge occurring during a normal activity cycle, based on the same FE model used in Sec. 3.3 (for details of FE setup and modeling, see Appendix. B). For each activity, we start by simulating the normal loading cycle, discretized into 61 quasi-static load steps. Based on the derived results, we identify the critical loading step at which peak stress occurs. We then generate a 13-step-width (approximately 20% of the cycle) Gaussian-modulated surge with $3\times$ peak overload, and simulate the surge-overload cycle with and without the hard stop activated. In short, we perform three simulations for each activity:

1) The normal load cycle.
2) The surge-overload cycle *without* the hard stop activated.
3) The surge-overload cycle *with* the hard stop activated.

For a concise presentation, we only show results for the three activities that cause the largest non-overload stress: 1) walking, 2) stair descent, and 3) jogging, as shown in Fig. 9 from top to bottom, respectively.

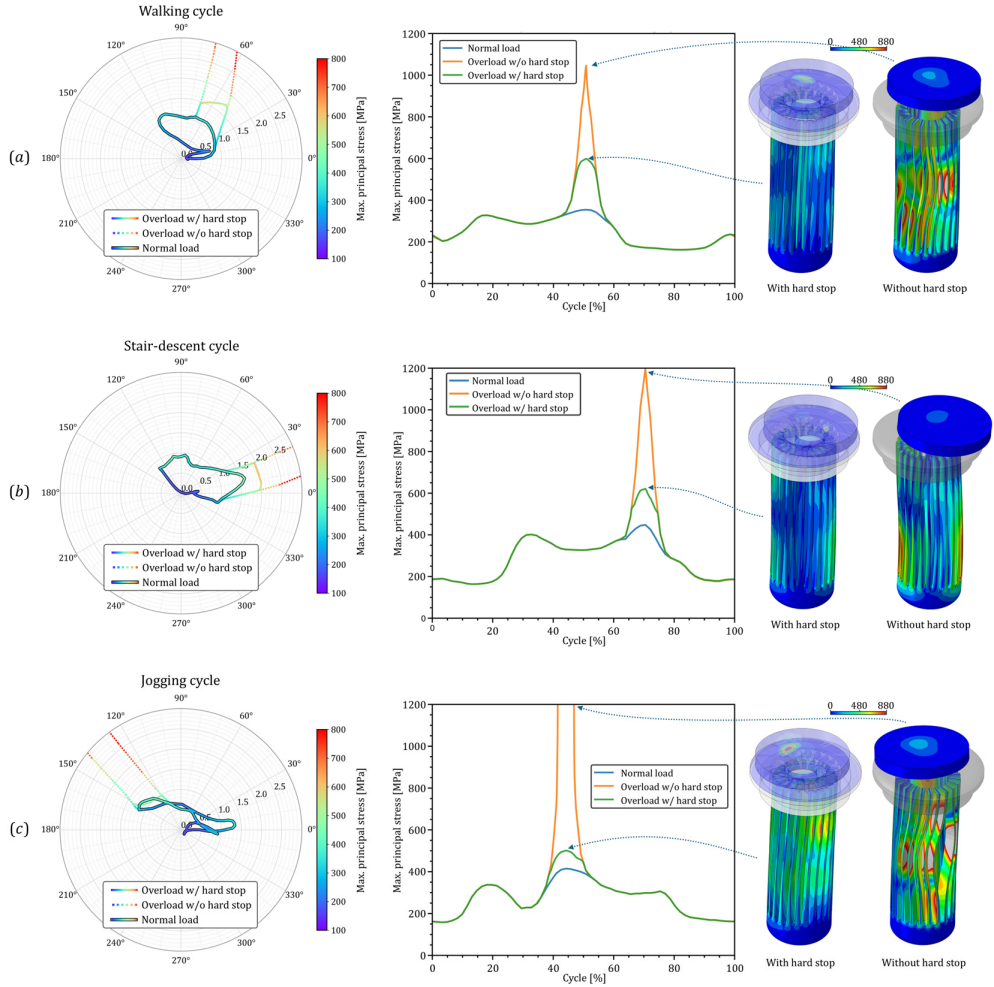

Fig. 9 Numerical validation of hard-stop performance during a) walking, b) stair descent, and c) jogging cycles under normal activity, $3\times$ overload surge with and without hard-stop protection.

Left column: translational motion curves of the stage hard-stop anchor $O_a$ from top-view projection (normal load cycle: solid curve with black stroke; overload without hard stop: dotted curve; overload with hard stop: solid curve without stroke).

Middle column: maximum principal stress as a function of activity cycle percentage (blue: normal load cycle; orange: overload without hard stop; green: overload with hard stop).

Right column: deformed mechanism in von Mises stress contours at peak stress step, with (left) and without (right) hard-stop protection.

The results in Fig. 9 show that during normal activities, the mechanism works with a maximum principal stress well below the fatigue limit of $480 MPa$. However, under a $3\times$ peak surge overload at the worst cycle percentage without hard-stop protection, a significant peak in stress is observed, exceeding both the fatigue and yield limits. Notably, we can observe from the deformed mechanisms that the blades begin to buckle under sufficiently high overloads. This is also evident in the translational motion plots in the left column of plots, where the absence of hard-stop protection leads to excessive translational deflection. From the stress-cycle curves in the jogging cycle (Fig. 9c), we observe that the mechanism completely loses stability under a $3\times$ peak overload. With hard-stop protection, the motion of the stage hard stop is promptly constrained upon engagement. This is followed by the stage hard stop rubbing against the ground hard stop as impact being absorbed, and an eventual disengagement as the overload is removed. For example, during the walking cycle (Fig. 9a), the overload surge begins at 41.6% of the cycle, with the hard stop engaging at 46.5%, just before the maximum stress in the mechanism surpasses $453MPa$. Upon reaching the peak overload at 50.8%, the maximum stress in the mechanism increases to its peak at $599.3MPa$. As overload is gradually removed, the hard stop disengages at 54.7% while the stress is reduced back to $469.1MPa$. Similarly, for the stair descent (Fig. 9b) cycle, the maximum stresses corresponding to hard-stop engagement, peak overload, and disengagement are $486.3MPa$, $621.3MPa$ and $498.8MPa$ respectively. For the jogging cycle (Fig. 9c), these critical values are $409.1MPa$, $501.3MPa$, and $453.5MPa$, respectively.

From these results, we find that the hard stop engages at stress levels between $410MPa$ and $490MPa$, which is effective but occurs slightly earlier than the ideal engagement stress ($480MPa$). This earlier engagement can be primarily attributed to the volume fraction $\phi_{hs}$ of $\mathcal{R}_{hs}$ being less than one. This is also visible in the results in Fig. 8c, where we observe certain blue areas (part of $\mathcal{R}_\sigma$) not covered by $\mathcal{R}_{hs}$. Motion toward these areas will result in an earlier engagement before the stress reaches the fatigue limit. Additionally, this behavior may be influenced by extra compliance in the system, including potential deformation in the connection piece linking the mechanism to the external case. Such deformation could transfer to the ground hard stop built upon the connection piece, further contributing to earlier engagement. Furthermore, we observe that the hard stop does not completely halt the increase in stress but instead rapidly makes it converge to a peak value between $500MPa \sim 620MPa$. This behavior is attributed to the inherent compliance of the hard-stop contact, as the contact pressure causes deformation of the polyethylene side of the hard stop. Nonetheless, under $3\times$ overload conditions, the stress remains well below the yield limit. If the overload is retrieved around $2.5\times$, the peak stress can safely stay below $480MPa$. This aligns with the design expectations, where the low frequency of extreme overload incidents is unlikely to significantly impact the mechanism's fatigue life.

In summary, the numerical validation results indicate that concerns regarding mechanism performance after hard-stop engagement and extra compliance of the system do not compromise the validity of the design. These factors are found to have a minimal impact on the major objective of fatigue life.

3.6  Experimental validation

In this section, we further validate our hard-stop design through experimental testing. Since obtaining the full stress field in a real compliant mechanism is not feasible, the primary objective of this experiment is to confirm the correct engagement of the hard stop and ensure that system compliance does not significantly affect the rigidity of hard-stop protection.

*Experimental setup*

We designed an experimental setup based on a modular version of the compliant intramedullary stem, as shown in Fig. 10a. This modular design ensures that all 22 caged-hinge blades and both hard-stop components can be easily replaced. Specifically, the intramedullary stem is assembled using a modular caged hinge and a central post connected via bolted joints on the bottom flange. The assembled stem is mounted on a four-post table, simulating its suspension within a rigid cage. The stem assembly (excluding the bolts and nuts) together with the stage hard stop is made from annealed TiAl64 Grade 5, while the ground hard stop is 3D printed with carbon-fiber-reinforced PLA.

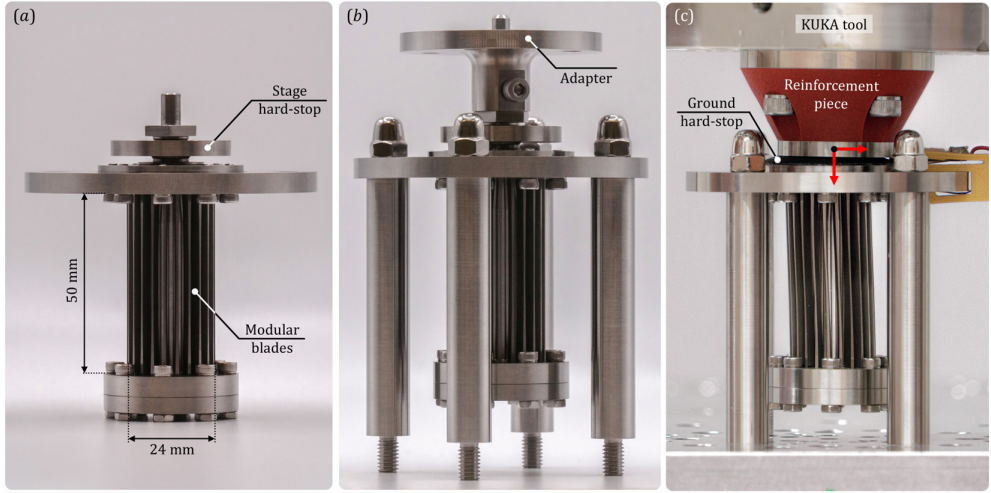

Fig. 10 Photos of the experimental setup: a) Modular intramedullary stem with a 22-blade modular caged hinge; b) Modular intramedullary stem mounted on the four-post fixture table and connected to the KUKA tool adapter; c) Experimental setup in operation under $500N$ shear force and $2000N$ compressive force.

Modularization of the mechanism introduces new geometries that make it impossible for the experimental assembly to exactly replicate the original mechanism's design. Therefore, the experimental assembly is designed to achieve similar mechanical behavior to that of the original mechanism (including variable thickness of the blades). As shown in Fig. 10a, the effective blade length is $50mm$, and the outer diameter of the blades is $24mm$. Consequently, the design parameters of hard stop are also slightly different from design I (Table 1), as detailed in Table 2.

Table 2 Geometric parameters of the hard-stop surfaces in experimental assembly.

| Stage hard stop | Ground hard stop | Vertical relations |
|---|---|---|
| $d_L^a = 12.0$ | $d_L^a = 12.0$ | $z_{ab} = 1.0$ |
| $d_S^a = 4.0$ | $d_S^a = 4.0$ | $z_{oa} = 3.0$ |
| $R_C^a = 11.0$ | $R_C^a = 25.89$ | |
| $\theta_O^a = 0°$ | $\theta_O^a = -10°$ | |
| $d_{hs}^a = 33.0$ | | |

The schematic setup of the experimental system is illustrated in Fig. 11. The green annular component represents the stage hard stop, which is clamped to the center post using a set nut, (Fig. 11b). The yellow annular component represents the ground hard stop, which is glued to the flange plate (blue part). External loads are applied via a KUKA KR 210 industrial robot arm, to which the center post is connected through an adapter (Fig. 11a). Additionally, an annular reinforcement piece is incorporated around the adapter connection area to minimize deformation in this region. For example, Fig. 10c shows the experimental system working under a $500N$ shear force and $2000N$ compressive force (after hard-stop engagement), where the red part is the reinforcement piece.

To achieve real-time detection of hard-stop engagement, we developed a novel sensor system using conductive coating on the ground hard stop. As shown in the upper part of Fig. 11c, the ground hard stop is coated with a thin layer of carbon-based conductive material (MG Chemicals 838AR). The conductive layer is connected to a voltage divider circuit (Fig. 11d), where the Titanium assembly (including the stage hard stop) is electrically grounded. Before contact engagement, the output end (marked blue) provides a high-voltage signal. Upon contact engagement between the stage hard stop and the conductive coating, an additional circuit loop is established, which divides more voltage out and results in a low voltage in signal output. The motion curve of the stage hard stop is primarily derived from tool motion data logged by the KUKA controller. However, under high loads, deformation of the adapter and drift of the fixture table introduce errors that cannot be detected by the KUKA robot alone. To address this, a computer vision system is employed to capture the absolute motion of the fixture and adapter, enabling calibration of the motion data recorded by the KUKA controller.

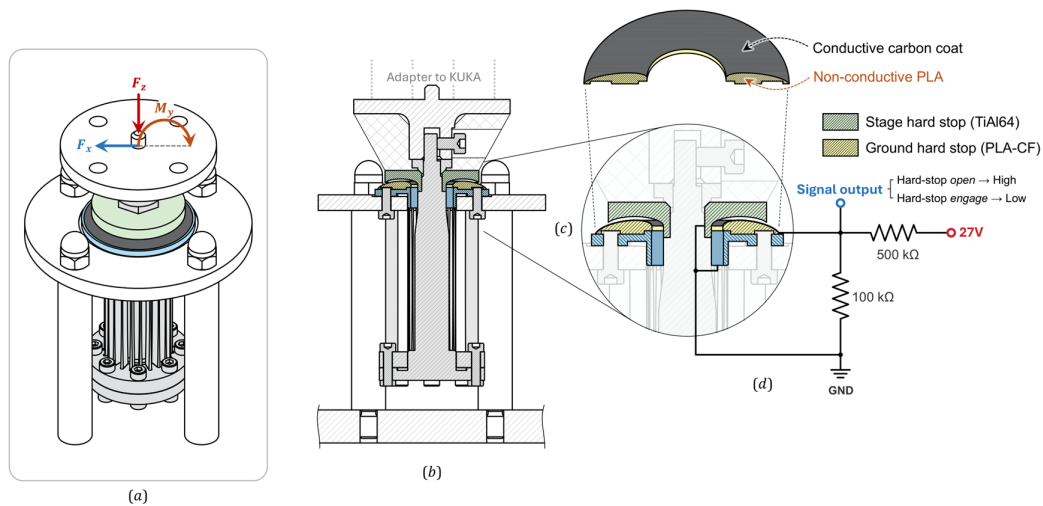

Fig. 11 Schematic of the experimental setup: a) Modular intramedullary stem mounted on a four-post fixture table with applied loads from the KUKA tool adapter; b) Sectional view of the experimental setup; c) Installation of the stage and grounded hard stops; d) Conductive coating-based sensor circuit for detecting contact engagement between hard-stop surfaces.

*Experimental results*

We conducted experimental validation of the modular intramedullary stem with the hard-stop design parameters listed in Table 2, performing two different tests: I) Bending with $M_y = 40Nm$ under a compressive force of $3000N$ (Fig. 12), and II) Shearing with $F_x = 500N$ under a vertical compressive force of $F_z = 2000N$ (Fig. 13). For both tests, no pre-rotation was applied, and $M_z$ was held at zero. In each figure, the upper plot illustrates the relationship between the applied load (shear force or bending moment) and the resulting translational and angular deflections, represented by the blue and orange curves, respectively. The lower part of each figure depicts the deformed assembly at various load levels.

For the shear force test shown in Fig. 12, we observe that as the shear force is applied, the stage hard stop undergoes significant translational motion with relatively small angular rotation. For example, as shown in Fig. 12-2, the stage hard stop experiences a translational deflection of $1.28mm$ and an angular deflection of $0.52°$ at $F_x = 151.95N$. At a high enough shear force of approximately $F_x = 277.76N$, contact engagement is detected as the sensor signal transitions from high to low (LED light switching from ON to OFF). Additionally, we can observe from the upper part of Fig. 12 that the motion curve experiences a significant slope change, implying the hard stop begins to absorb the load. However, we also find that, even at very low rates, the stage hard stop continues to move slightly after the engagement. For example between Fig. 12-4 and Fig. 12-6, the stage hard stop experiences a further translational motion of $0.11mm$ and an angular rotation of $0.13°$ as the shear force increases from $277.76N$ to $491.43N$. We consider this after-engagement motion as a result of the relatively low stiffness and surface quality of the 3D-printed ground hard stop. While the 3D-printed hard stop is ideal for rapid prototyping, shape optimization, and experimental validation, the final product will use surface-hardened UHMWPE for improved performance. Moreover, the experimental ground hard stop is a relatively thick annular piece, whereas the final product will incorporate a thin UHMWPE layer over a titanium base.

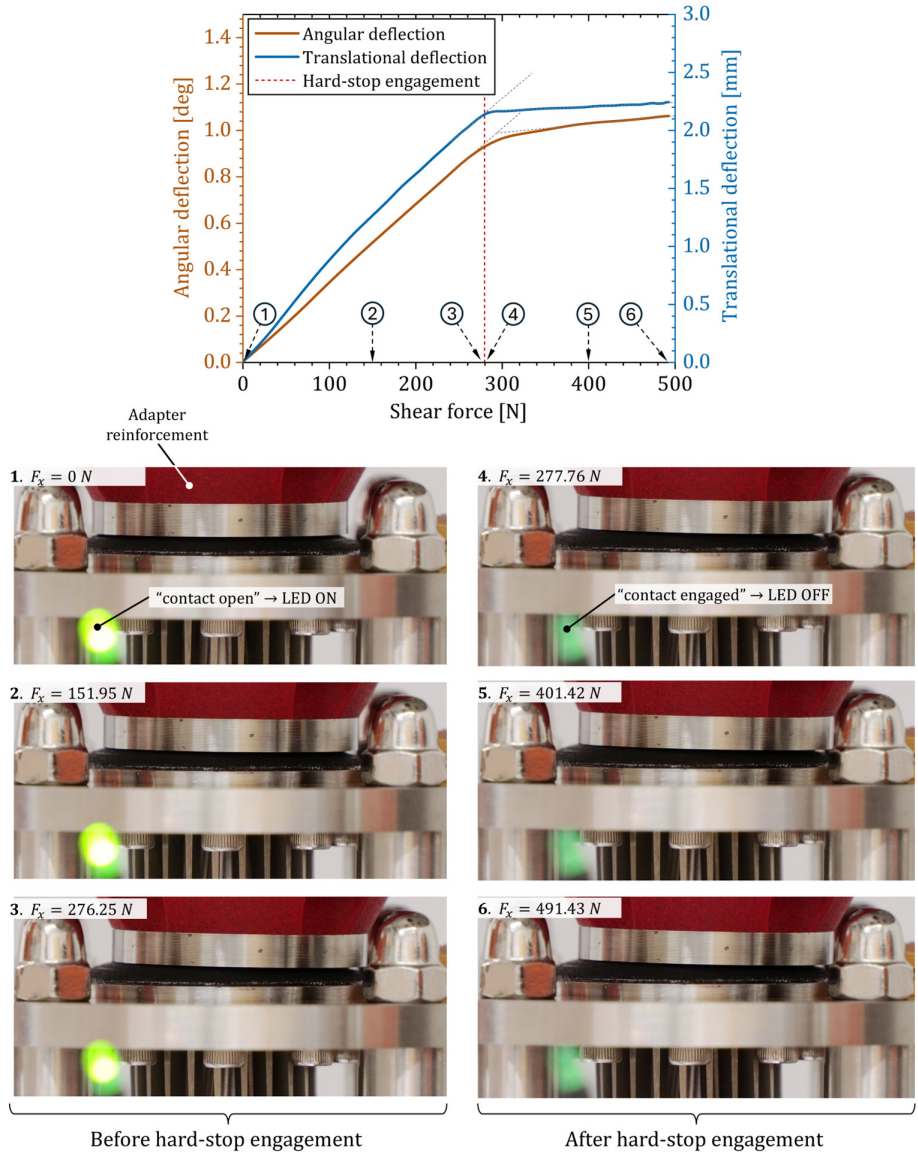

Fig. 12 Experimental results of hard-stop performance with application of $F_x = 500N$ shear force and $F_z = 2000N$ compressive force. Hard-stop engagement indicated by the LED light switching from ON (contact open) to OFF (hard stop engaged).

The bending test with $M_y = 40Nm$ and $F_z = 3000N$ is shown in Fig. 13. The results demonstrate a similar transition in performance after engagement to that observed in the shear test. However, the primary difference is that in the bending test, the stage hard stop experiences a relatively larger angular rotation and smaller translational motion. For example, from the unloaded state (Fig. 13-1) to the near-contact state of hard stop (Fig. 13-3), the stage hard stop experiences a translational deflection of $0.43mm$ and an angular deflection of $2.18°$. This results in the engagement of the hard stop occurring at the outer edge of the stage hard stop (Fig. 13-4, 5, 6). Additionally, we observe that the translational motion of the stage hard stop is almost halted after the contact engagement (which only increased $0.02mm$ from contact engagement to $M_y = 40Nm$), while the angular rotation still undergoes a transition phase before stabilizing to a near-zero increase in slope. Specifically, the stage hard stop rotates an additional $0.41°$ after engagement before the contact becomes fully stable. We notice that this after-engagement rotation is attributed to the sharp edges of the fabricated stage hard stop. The concentration of contact pressure at these sharp edges causes further deformation in the angular direction but halts further translational motion. This also explains why we did not see further translational sliding of the stage hard stop. These findings imply the importance of adding fillets to the outer edge of the stage hard stop in the final product to reduce contact pressure concentrations.

The experimental results presented in this section validate proper engagement of the hard stop, which successfully absorbs overloads and protects the compliant mechanism. However, additional compliance resulting from the deformation of the ground hard stop was observed, primarily leading to limited contact stiffness and localized pressure concentration. To address these issues in real product manufacturing, using harder materials for the hard stop, reducing the thickness of the hard-stop layer, and incorporating fillet transitions on the hard-stop edges are recommended. These improvements can effectively minimize after-contact motion and enhance overall performance.

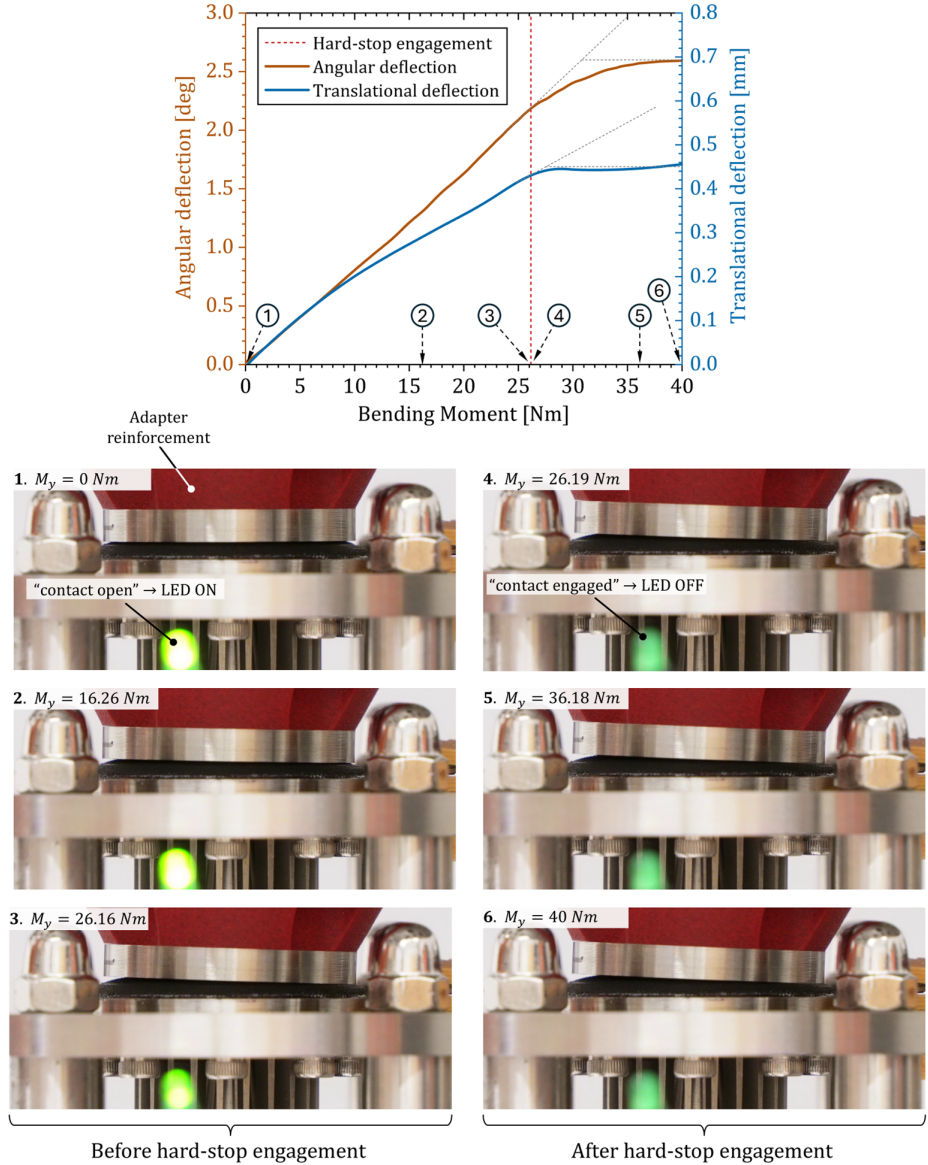

Fig. 13 Experimental results of hard-stop performance with application of $M = 40 Nm$ bending moment and $F_z = 3000N$ compressive force. Hard-stop engagement indicated by the LED light switching from ON (contact open) to OFF (hard stop engaged).

## 4. Conclusion

In this study, we proposed a general design framework for synthesizing multi-DOF hard-stops for compliant mechanisms. We first framed the contact-based motion limit as a multi-dimensional manifold embedding: the hard-stop-free space inscribed within the mechanism's safe workspace. Under highly uncertain overloads, we found that long-term hard-stop performance is a precise balance between ensuring sufficient protection and minimizing overprotection, formulated as maximizing the volume fraction of the hard-stop-free workspace within the safe workspace. Our theoretical analysis shows that combinations of independent single-DOF hard stops inevitably lead to overprotection, due to their inability to couple motion limits across dimensions. In contrast, our general multi-DOF hard-stop solution couples motion limits between DOFs via radial constraints, which are implemented through optimized 3D surface-to-surface contacts. We introduced a complete design framework for these multi-DOF hard stops that incorporates: 1) analysis of the stress response function $\mathcal{R}_\sigma$, 2) modeling of hard-stop shape functions $\mathbf{\Omega}_a$ and $\mathbf{\Omega}_b$, 3) shape optimization based on maximizing the volume fraction of the hard-stop-free workspace.

To demonstrate this design framework, we conducted a representative case study on a compliant intramedullary stem for use in TKA. We first used FEA to map the mechanism's stress response, which revealed an inherently skewed safe stress space in the workspace defined by translational and angular deflections. We attribute this skew to the synergistic and antagonistic interactions between these motions, and note that skew is a common feature of two-DOF stress responses. To maximize motion within the safe workspace, we created an oblique elliptic torus hard-stop shape that yields a skewed hard-stop-free workspace and can be optimized with manageable parameters. Finally, through parametric optimization, we generated an optimal hard-stop design that achieves the following objectives: 1) zero contact during normal knee activities, 2) maximum principal stress below the fatigue limit during normal load to moderate overloads, and 3) maximum von Mises stress below the yield limit during low frequency, high-level overloads. The performance of the design was validated through numerical simulations and experimental tests. The numerical results confirmed that the hard stop successfully protects the stress surges under a $3\times$ overload. The experimental results demonstrated proper engagement and load absorption of the hard stop under $2\times$ overloads. These tests also revealed the influence of extra compliance—primarily due to contact stiffness and pressure concentration—on hard-stop performance, providing guidance for future design improvements in real-world applications.

This research establishes a foundational framework for precisely modeling and designing hard stops in compliant mechanisms, applicable to a broad range of compliant systems. Future work will primarily focus on understanding the after-contact behavior of hard stops, with particular emphasis on tribological effects, which are critical for enhancing the efficiency and durability of hard stops.

## 5. Acknowledgments

The research is supported by the NIH New Innovator Award.

## Data Availability

Data will be made available on request.

## Appendix. A  Stress response of the cylindrical compliant system

In Sec.3.3 and Fig. 4, we have demonstrated that the caged-hinge mechanism, inverted via a flanged post, exhibits a "skewed" stress response to the input motion $U$ defined by the angular and translational deflections. This characteristic stress response, as further realized, is not exclusive to the representative mechanisms. Instead, it represents a more general phenomenon in circular symmetric mechanisms. This appendix provides a conceptual explanation of why a "skewed" stress response is a common feature in similar circular symmetric mechanisms.

We begin with an intuitive explanation for this phenomenon. When angular and translational deflections align in the same direction, the input can be reduced to a single-direction load. However, if the angular and translational deflections oppose each other, the input load must necessarily include a combination of high forces and moments. For example, as illustrated in Fig. 14a, we consider the positive (+) direction as rightward and the negative (−) direction as leftward. A single lateral shear force $F$ applied in the (+) direction will result in angular and translational deflections both in the (+) direction, denoted as $\vartheta_a(+)$ and $\delta_a(+)$. Conversely, as shown in Fig. 14b, if a combination of $\vartheta_a(-)$ and $\delta_a(+)$ are given, as a result, the load condition causing these deformations requires a combination of significantly higher shear force $F'(+) > F(+)$ and a bending moment $M(-)$ in the opposite direction. This result arises because $M(-)$ must generate $\vartheta'_a(-)$ much larger than $\vartheta_a(+)$ to "flip" the resultant deformation to the negative direction. Similarly, as $M(-)$ also induces a translational deflection in the (−) direction, to maintain a translational deflection in the (+) direction, the shear force $F'$ must be much larger than $F$.

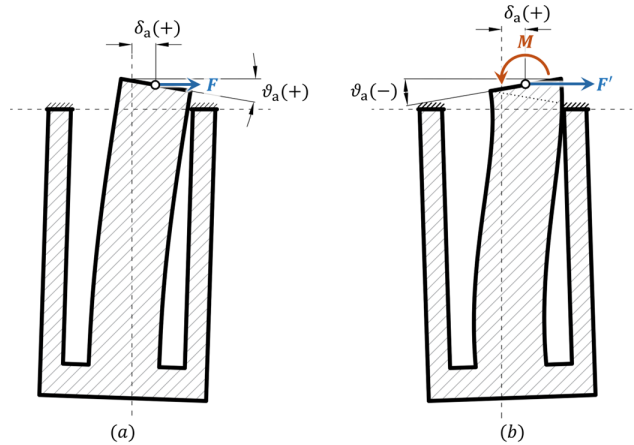

Fig. 14 Illustration of potential loading conditions with angular and translational deflections in the same (a) or opposing (b) directions

Using the same method described in Sec. 3.3 and Sec.3.5, we validated this stress response via a numerical simulation of a simple cantilever beam with a circular cross-section subjected to combined loading conditions $U|_{F_z=3kN} = (\delta_a, \vartheta_a, \theta_{sep})$, as shown in Fig. 15. We can observe that, for $\theta_{sep} \neq 90°$, the stress response exhibits a "skewed" distribution. For example, at $\theta_{sep} = 0°/180°$, a "low stress" region is observed in the first and third quadrants, corresponding to the antagonistic effects of $\vartheta_a$ and $\delta_a$. While $\vartheta_a$ and $\delta_a$ are theoretically capable of canceling each other, regardless of the amplitudes of $\vartheta_a$ and $\delta_a$, we did not observe an infinitely large $\mathcal{R}_\sigma$. This is attributed to the absolute shear deformation of the mechanism. In summary, a skewed stress response is a general feature of slender bodies of revolution, extending beyond the caged-hinge mechanisms that are the primary focus of this work.

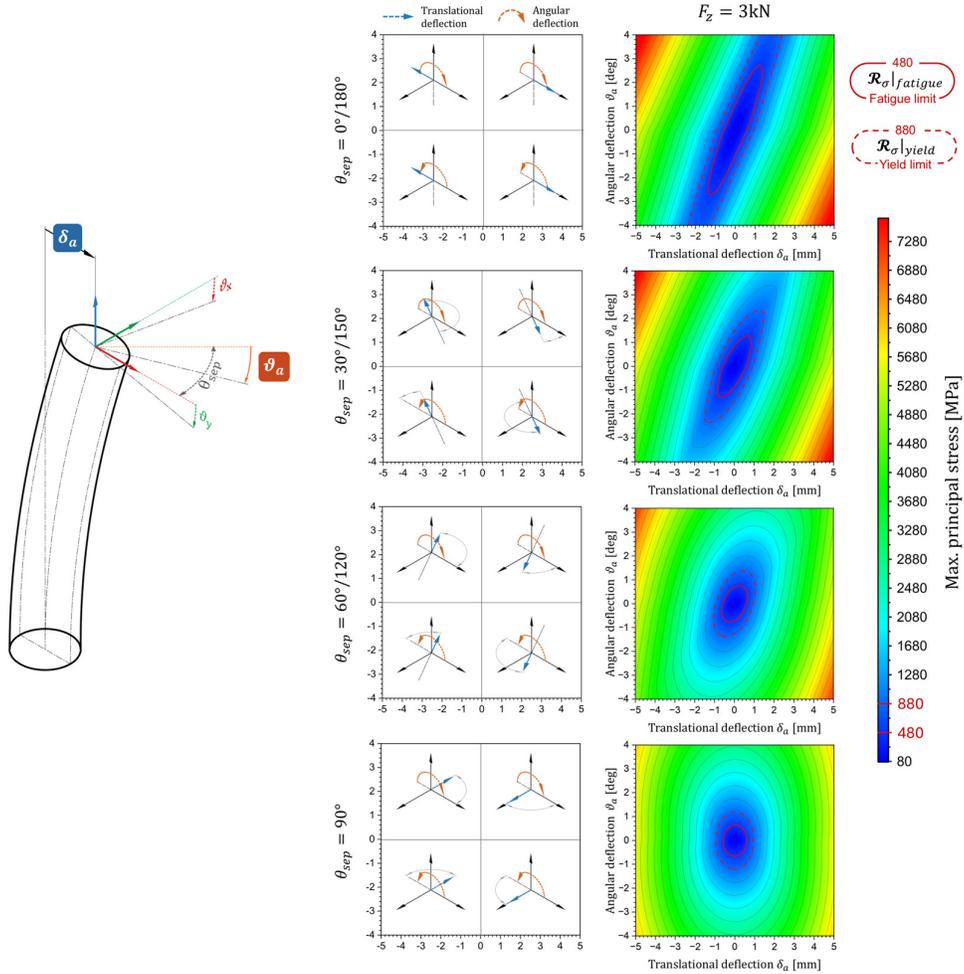

Fig. 15 Heatmaps of maximum principal stress in a simple cylinder cantilever beam as functions of translational and angular deflections, shown for separation angles of 0°/180°, 30°/150°, 60°/120°, and 90° (from top to bottom) under a compressive force of $3kN$.

## Appendix. B  Finite element modeling details

The numerical analysis in this study, including the stress response analysis presented in Sec. 3.3 and the numerical validation conducted in Sec. 3.5, was conducted using a finite element (FE) model developed in ABAQUS 2023. The FE model was constructed as a single body piece comprising components with different materials, as shown in Fig. 16a. The three components are shown in the cross-sectional view in Fig. 16b:

    I) Compliant intramedullary stem: constructed with a caged hinge inverted by a flanged post, represented by the grey elements.

    II) The grounded case supporting the compliant stem, shown as the green elements.

    III) The ground hard-stop layer, illustrated by yellow elements in Fig. 16b.

The stage and ground hard stops are modeled using the geometrical parameters listed in Table 1, as shown in Fig. 16c.

*Materials*

Part I and II are modeled as isotopic elastic materials with the mechanical properties of Ti-Al-4V. The Young' Modulus is set to be $104.8 GPa$, and the Poisson's ratio is $0.31$. Part II (ground hard stop) is modeled as isotropic elastic material with Young' Modulus of $799.5 MPa$ and Poisson's ratio of $0.46$ [58]. Viscoelasticity and other time-dependent properties are not considered in our FE model.

*Loads and boundary conditions*

The input load vector $W$ is applied at the load reference point $O_L$, which is kinematically coupled to the top surface of the flanged post (represented by the yellow face in Fig. 16b). A driven reference point $O_a$ is defined to represent the motion anchor of the stage hard stop, which derives the motion vector $U$ of the stage hard stop. A vertical offset between $O_a$ and $O_L$ is set as $Z_{OL} = 9mm$, representing accounting for the thickness of the titanium stage hard stop ($3\ mm$) and the typical thickness ($6\ mm$) of the TKA polyethylene insert [59]. The bottom-outer surface of Part III is fully constrained using an encastre boundary condition. Moreover, a surface-surface contact pair is established between the stage (primary surface) and the ground (secondary surface) hard stops. The normal behavior of the contract pair is set as rigid, while the tangential behavior is defined as an isotopic frictional sliding with a coefficient of friction of $0.1$.

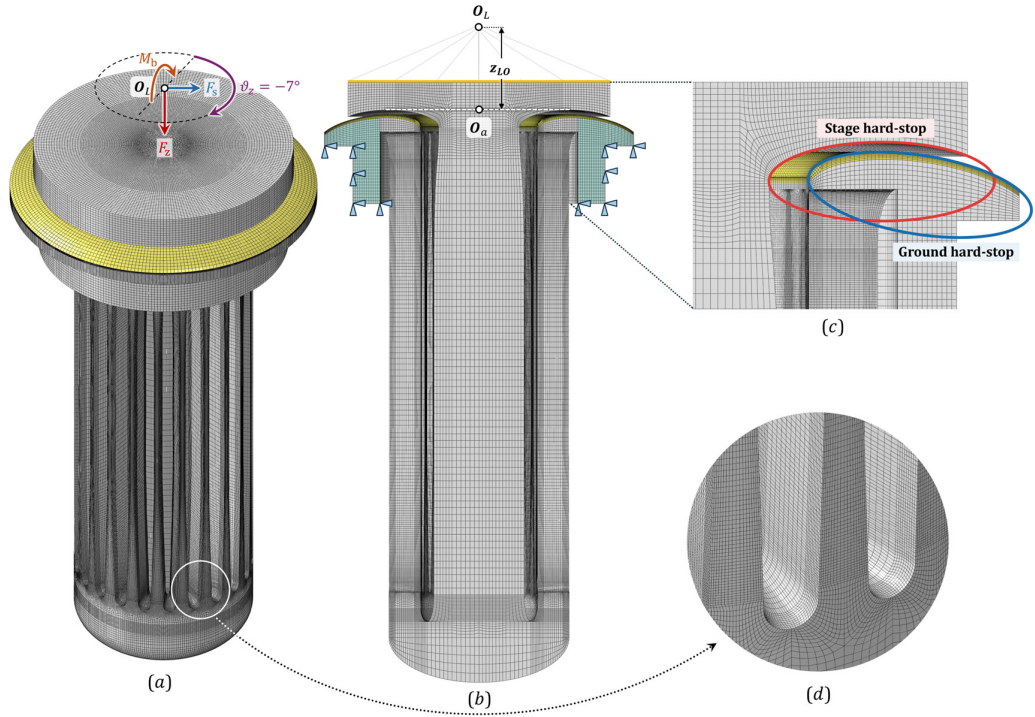

Fig. 16 Finite element modeling of the compliant intramedullary stem with hard stops: a) Full view of the FE model with illustration of applied loads; b) Cross-sectional view of the FE model highlighting different components; c) Zoomed-in view of the hard stops; d) Zoomed-in view of the refined mesh in areas with potential stress concentrations.

*Meshing*

The geometric model was discretized using a fully structured (mapped) mesh with C3D20R quadratic hexahedral elements in ABAQUS. Specific regions of potential stress concentration, such as the blade fillets shown Fig. 16d, were seeded with a finer mesh to address stress singularities and accurately resolve the localized stress fields.

To determine the appropriate mesh density, a convergence study was conducted on various mesh refinement levels. For each mesh level, a load of $[F_x, F_y, F_z, M_x, M_y] = [-6.84N, 388.48N, -2286.8N, -6.45Nm, -11.15Nm]$ with $\theta_z = -7°$ is applied, and the resulting maximum stresses are then evaluated. The dependence of maximum stress values on mesh refinement levels is shown in Fig. 17. The results demonstrate that as the mesh density increases beyond Level 4, the maximum stress values converge, showing minimal variation with further increases in the total degrees of freedom. This indicates that the computational accuracy of the solution is no longer sensitive to mesh size at level 4 and higher refinement. Therefore, Level 4 mesh density was selected for all our numerical simulations, yielding a model with approximately 50 million degrees of freedom.

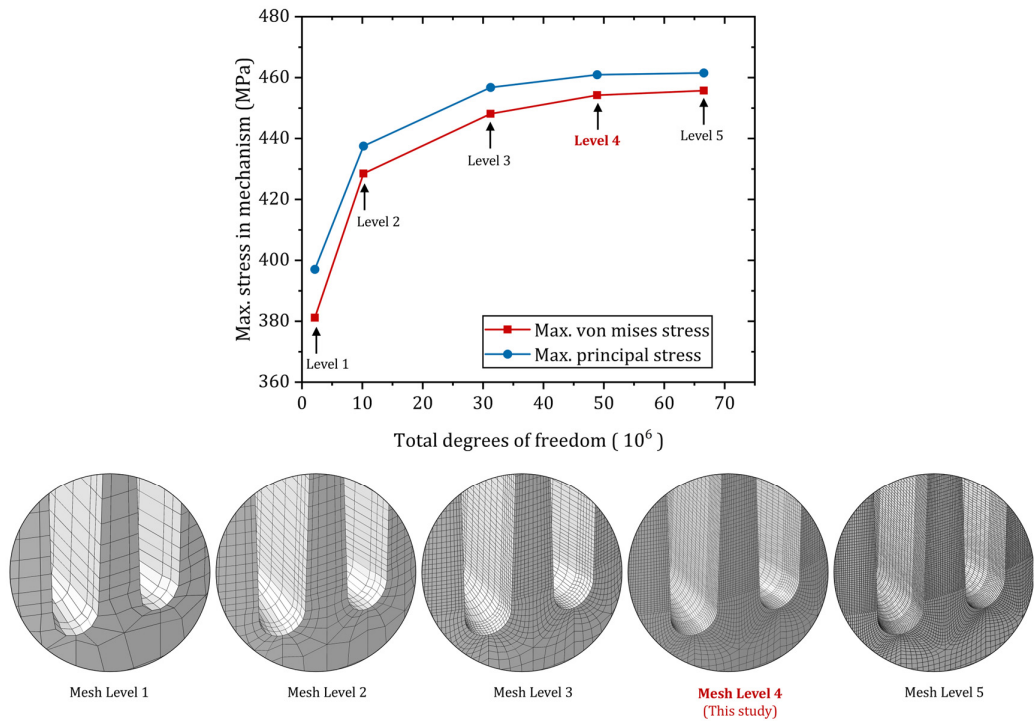

Fig. 17 The variation of maximum principal stress and von Mises stress as a function of the total degrees of freedom for the FE model, evaluated across mesh density levels from 1 to 5.